\documentclass[lettersize,journal]{IEEEtran}
\usepackage{amsmath,amsfonts}
\usepackage{algorithm}
\usepackage{array}
\usepackage[caption=false,font=normalsize,labelfont=sf,textfont=sf]{subfig}
\usepackage{textcomp}
\usepackage{stfloats}
\usepackage{url}
\usepackage{verbatim}
\usepackage{graphicx}
\usepackage{cite}
\usepackage[utf8]{inputenc}
\usepackage{hyperref}
\usepackage{algpseudocode}
\usepackage{xcolor}
\graphicspath{{Figures/}{../Figures/}}
\usepackage[caption=false]{subfig}
\usepackage{tabularx}
\usepackage{float}
\usepackage{multirow}
\usepackage{amsmath}
\usepackage{enumerate}
\usepackage{hyperref}
\usepackage{graphicx} 
\usepackage{amsthm,amssymb}
\usepackage{bm}
\usepackage{hyperref}
\usepackage{dsfont}
\usepackage[caption=false]{subfig}
\usepackage{tabularx}
\usepackage{float}
\usepackage{algpseudocode}
\usepackage{multirow}
\usepackage{titling}

\pretitle{\begin{center}\LARGE}
\posttitle{\par\end{center}\vskip 0.5em}
\preauthor{\begin{center}\large \lineskip 0.5em}
\postauthor{\par\end{center}}
\predate{\begin{center}\large}
\postdate{\par\end{center}}

\begin{document}
\title{An Interpretable Approach to Load Profile Forecasting in Power Grids using Galerkin-Approximated Koopman Pseudospectra}

\author{Ali Tavasoli, Behnaz Moradijamei, Heman Shakeri 
\thanks{Ali Tavasoli is with the School of Data Science, University of Virginia, Charlottesville, VA, USA. (at9kf@virginia.edu)}
\thanks{Behnaz Moradijamei is with the Department of Mathematics \& Statistics, James Madison University, Harrisonburg, VA, USA.}
\thanks{Heman Shakeri is with the School of Data Science, University of Virginia, Charlottesville, VA, USA. (hs9hd@virginia.edu)}}


\date{}
\maketitle
\vspace{-2em}  
    \textit{This work has been submitted to the IEEE for possible publication. Copyright may be transferred without notice, after which this version may no longer be accessible.}
\vspace{1em}

\begin{abstract} This paper presents an interpretable machine learning approach that characterizes load dynamics within an operator-theoretic framework for electricity load forecasting in power grids. We represent the dynamics of load data using the Koopman operator, which provides a linear, infinite-dimensional representation of the nonlinear dynamics, and approximate a finite version that remains robust against spectral pollutions due to truncation.
By computing $\epsilon$-approximate Koopman eigenfunctions using dynamics-adapted kernels in delay coordinates, we decompose the load dynamics into coherent spatiotemporal patterns that evolve quasi-independently. Our approach captures temporal coherent patterns due to seasonal changes and finer time scales, such as time of day and day of the week. This method allows for a more nuanced understanding of the complex interactions within power grids and their response to various exogenous factors.
We assess our method using a large-scale dataset from a renewable power system in the continental European electricity system. The results indicate that our Koopman-based method surpasses a separately optimized deep learning (LSTM) architecture in both accuracy and computational efficiency, while providing deeper insights into the underlying dynamics of the power grid\footnote{The code is available at \href{https://github.com/Shakeri-Lab/Power-Grids}{github.com/Shakeri-Lab/Power-Grids}.}.
\end{abstract}

 \begin{IEEEkeywords}
Power grids, load forecasting, Koopman operator, $\epsilon$-spectral analysis, spatiotemporal patterns, clustering.
\end{IEEEkeywords}
	
\section{Introduction}

\IEEEPARstart{E}{lectricity} load forecasting plays a crucial role in power grid optimization and management. Current approaches to load forecasting can be broadly categorized into conventional statistical methods and machine learning methods. Conventional statistical models use time series analysis to extract information and predict future points based on historical data. Examples include ordinary regression models \cite{Weron2006book}, auto-regressive moving average model (ARMA) \cite{Pappas2010ARMA}, Autoregressive integrated moving average model (ARIMA) \cite{Chodakowska2021ARIMA}, and exponential smoothing \cite{Taylor2007}.
To better handle the nonlinear, highly volatile, and uncertain features of load profiles, methods such as random forest (RF) \cite{Dudek2015RF}, support Vector Regression (SVR) \cite{Chen2017SVR}, gradient boost \cite{Patnaik2021XGBoost}, Gaussian process \cite{Shepero2018GP}, or fuzzy systems \cite{Efendi2015fuzzy} have been developed. These approaches often deliver improved forecasting accuracy, albeit at the cost of reduced interpretability. More recently, deep learning techniques such as Deep Neural Networks (DNNs) \cite{He2024NN}, Convolutional Neural Networks (CNNs) \cite{Marcus2018CNN}, and Long Short-Term Memory (LSTM) \cite{Kong2019LSTM} have emerged. These methods often outperform conventional statistical models due to their flexibility in handling complex features of load profiles \cite{Nti2020review}.

\IEEEpubidadjcol

However, while these methods have shown success in various applications, they often struggle with the complex, nonlinear, and high-dimensional nature of power grid dynamics. Deep architectures are prone to overfitting, which hinders their forecasting accuracy. Moreover, most neural network approaches lack interpretability, making them difficult to analyze and interpret \cite{Brunton&Kutz2022}. They often struggle with extrapolation and generalization beyond the training data, limiting their application in physical and engineering systems where models must be valid beyond the training range for prediction, optimization, and control purposes.


In this paper, we present an alternative approach to load forecasting that leverages an operator-theoretic framework. Instead of attempting to predict exact trajectories in the complex power system, we focus on the probabilistic behavior of the system. This perspective allows us to capture the inherent uncertainties and nonlinearities present in the interconnected power grid dynamics. Following the formulation offered by Mezić \cite{Mezic2005Spectral}, we employ the Koopman operator, which provides a linear, infinite-dimensional representation of nonlinear dynamics, allowing us to study the evolution of observables rather than state variables.
This approach connects with physically-interpretable data-driven methods that extract and decompose the intrinsic manifold of the data, producing coherent patterns of dynamics equivalent to the Koopman operator's eigenfunctions \cite{Mezic2005Spectral}. In signal processing, it can be viewed as an extension of Fourier analysis, replacing predefined frequency bases with intrinsic spectra and temporal patterns with spatiotemporal patterns \cite{Brunton2021Fourier}. Moreover, the associated timescales with the eigenfunctions facilitates the analysis of multiscale systems like interconnected power grids \cite{Giannakis2019Spectral}.

Various approaches have been proposed to approximate a truncated version of the Koopman operator from data, including Dynamic Mode Decomposition (DMD) \cite{Rowley2009, Schmid2010DMD, Tu2014DMD}, Extended DMD \cite{Williams2015EDMD}, and other methods \cite{Fjii2019GraphDMD, Mezic2017Ergodic, Mezic2020Spectral, Giannakis2019Galerkin, Lusch2018DL}. While DMD has shown promise in power grid analysis, demonstrating improved accuracy and adaptability to exogenous factors \cite{Mohan2018DMD, Kong2020DMD, Kutz2022GP}, it still faces limitations in capturing complex nonlinear and transient phenomena \cite{Schmid2022DMD}.
These operator-based methods offer significant advantages over traditional spectral analysis techniques such as Fourier analysis \cite{Zhong2015DFT}, wavelet analysis \cite{Li2017wavelet}, and PCA \cite{Rydin2020PCA}, which have been widely used in power grid studies \cite{Zhong2012frq}. While these conventional methods have their merits, they often fail to capture multi-scale frequencies and are limited to linear regions of the state space \cite{Brunton2021Fourier, Giannakis2012NLSA}. Crucially, they struggle to provide information about chaotic dynamics and continuous spectra. This limitation is particularly critical for power grid applications due to the complex nonlinearities resulting from the network effects, leading to different transient behaviors that propagate through the grid. Understanding these dynamics is crucial for comprehending power grid behavior \cite{Yu2003chaos, Halekotte2021chaos}.

Our proposed approach addresses the limitations of traditional methods by employing a framework based on ergodic theory \cite{Mezic2005Spectral, Giannakis2019Delay} that allows us to leverage powerful mathematical to extract spatiotemporal load dynamics patterns in complex dynamical systems.  We compute $\epsilon$-approximate Koopman eigenfunctions using dynamics-adapted kernels in delay coordinates and leverage integral kernel operators to approximate Laplace-Beltrami operator eigenfunctions, effectively capturing the geometry of the nonlinear data manifold \cite{Coifman2006Diffusion, Giannakis2012NLSA}. This approach allows us to obtain a decomposition of the load dynamics into coherent spatiotemporal patterns that are both intrinsic and robust.
The resulting patterns evolve quasi-independently according to their respective frequencies, enabling predictability based on linear dynamics and encoding rich dynamical features at multiple time scales. This approach is particularly well-suited for analyzing the complex, nonlinear dynamics of power grids, as it can capture the chaotic and continuous spectra that are crucial for understanding system behavior.
To enhance the interpretability of our results, we complement our Koopman analysis with a clustering approach based on another kernel method \cite{PHATE2019}. This additional step helps to identify and group power stations with similar dynamical properties, providing a more nuanced understanding of the system's behavior. By capturing both local and global nonlinear data structures, this approach effectively reveals stations with synchronized dynamics, allowing for a more comprehensive analysis of the power grid's behavior while maintaining the integrity of our Koopman-based method.
Our approach contributes to the field by:
\begin{itemize}
    \item Providing a robust method for capturing complex, nonlinear dynamics in power grids
    \item Enabling multi-scale analysis of load dynamics
    \item Improving predictability through the use of linear dynamics in the $\epsilon$-eigenfunction space
    \item  Enhancing interpretability through the combination of Koopman eigenfunctions and clustering techniques
\end{itemize}

The rest of this paper is organized as follows. Section \ref{sec:Koopman} provides a detailed introduction to the Koopman operator and its mathematical foundations. In Section \ref{sec:data-driven}, we describe our data-driven approach for identifying the Koopman operator and its eigenfunctions. Section \ref{sec:cluster} introduces the clustering method used to group power stations with similar dynamical properties. Section \ref{sec:Results} presents our results, including the application of our method to a large-scale dataset from the continental European electricity system, and compares our approach with traditional and deep learning methods. Finally, Section \ref{sec:Conclusion} offers concluding remarks and discusses potential future directions for this research.

\section{Operator-Theoretic Framework}\label{sec:Koopman}
The analysis of complex dynamical systems, such as power grids, often involves interconnected multiscale nonlinear dynamics that are challenging to study directly. To overcome these difficulties, we adopt an operator-theoretic framework, shifting our focus from exact trajectory prediction to a probabilistic description of the system's behavior.
Let us consider a dynamical system with state space $\mathbb{X}$ and a flow map $\Phi_\tau : \mathbb{X} \rightarrow \mathbb{X}$, where $\Phi_\tau (\textbf{X}_t) = \textbf{X}_{t+\tau}$ for $\tau \geq 0$. Instead of analyzing $\Phi_\tau$ directly, we opt to study the probability that trajectories belong to subsets $\mathbb{A} \subset \mathbb{X}$.
Central to this framework is the transition density function $p_\tau:\mathbb{X} \times \mathbb{X} \rightarrow [0,\infty)$ of $\Phi_\tau$, defined as:
\begin{equation}
\mathbb P[\Phi_\tau (\textbf X_t)\in\mathbb A\mid\textbf X_t=x]=\int_{\mathbb A} p_\tau(x,y)\mu(dy)
\end{equation}
where $\mu$ is a measure on $\mathbb{X}$. This approach allows for the use of linear operators to analyze the probabilistic behavior of the system. Two key operators in this framework are the Perron-Frobenius operator, which describes how probability densities evolve, and its adjoint, the Koopman operator. While both operators provide valuable insights, we will focus on the Koopman operator due to its particular advantages in analyzing observables of the system.
The Koopman operator, $\mathcal{K}_\tau$, acts on observables (functions of the state) rather than on the state space itself. In the context of power grids, the recorded signal at $d$ sensors is viewed as an observation function $F:\mathcal X\mapsto\mathbb R^d$.
The dynamical system $(\mathcal X , \Phi^t)$ is assumed to possess ergodic measures; hence there exists a probability measure $\mu$ on $\mathcal X$, invariant under the flow map $\Phi^t$, such that for every integrable function $f : \mathcal X\mapsto\mathbb C$, the time average $\bar f$ of $f$ converges to the expectation value $\bar f = \int_\mathcal X f d\mu$.

Associated with the triplet $(\mathcal X , \Phi^t , \mu)$, we consider a Hilbert space $\mathcal H = L^2(\mathcal X , \mu)$ of square-integrable observables with respect to $\mu$. The group of unitary Koopman operators $U^t:\mathcal H\mapsto\mathcal H$ governs the evolution of observables under $\Phi^t$. Specifically, given $f\in\mathcal H$, $g=U^tf$ is defined as the observable satisfying $g(x)=f(\Phi^t(x))$ for $x\in \mathcal X$.
Our approach is to identify dynamically intrinsic temporal patterns through eigenfunctions of the Koopman operator. This will lead to a decomposition of the observation map $F$ into a corresponding set of spatiotemporal load patterns.
An observable $\psi_j\in\mathcal H$ is a Koopman eigenfunction if it satisfies the eigenvalue equation
\begin{equation}\label{eq:eigen}
U^t\psi_j=e^{i\omega_jt}\psi_j 
\end{equation}
for all $t \in \mathbb R$. The eigenfrequency $\omega_j$ is a real-valued frequency associated with the eigenfunction $\psi_j$. In measure-preserving dynamical systems, the Koopman eigenvalues remain on the unit circle in the complex plane, and the corresponding eigenfunctions evolve periodically under the dynamics. This is the key to the predictability of coherent patterns of dynamics. The Koopman eigenvalues and eigenfunctions appear as complex-conjugate pairs, and the Koopman eigenfunctions that correspond to different eigenfrequencies are orthogonal in the Hilbert space $\mathcal H$.

Let $\mathcal D$ be the space of Koopman eigenfunctions that is invariant under $U^t$, and is the closure of the span of ${\psi_j}$. Every $f \in \mathcal D$ can be decomposed as $f=\sum_j\hat f_j\psi_j$, where $\hat f_j=\langle f,\psi_j\rangle$ is the inner product in $\mathcal H$. The norm in $\mathcal D$ is denoted by $\parallel \cdot \parallel$.
Moreover, the dynamical evolution of $f$ can be computed in closed form via
\begin{equation}\label{eq:evolution} 
U^tf = \sum_j\hat f_je^{i\omega_jt}\psi_j 
\end{equation}
For every continuous flow $\Phi^t$, the family of operators $U^t$ has a generator $V$, which is a skew-adjoint operator, defined as
\begin{equation}\label{eq:Generator} 
Vf:=\lim_{t\to 0}\frac{1}{t}(U^tf-f), \quad f\in D(V)\subset L^2(\mathcal X,\mu) 
\end{equation}
Operators $U^t$ and $V$ have simple eigenvalues and share the same eigenfunctions.

\section{The Data-Driven Approach}\label{sec:data-driven}
To put the above approach in a numerical setting, we start with a collection of $N$ samples $F(x_1),...,F(x_N)$, organized in a time-ordered manner, where each $F(x_i)\in \mathbb{R}^d$. The value of $x_n$ is determined by the function $\Phi^{n\Delta t}(x_0)$, where $\Delta t$ is the sampling interval.
We construct a delay coordinate map from $F$ by embedding $\mathcal X$ in a manifold in $\mathbb R^{Qd}$ as:
\begin{equation}\label{eq:delay}
F_Q(x) = (F(x),F(\Phi^{-\Delta t}x), \cdots , F(\Phi^{-(Q-1)\Delta t}(x)))
\end{equation}
where $Q$ (an integer) is the number of delays.
Next, we define a kernel function $k_Q : \mathcal X\times \mathcal X\mapsto \mathbb R_+$ to measure the similarity of points in $\mathcal X$ based on the observation function $F_Q$. We use the radial Gaussian kernel with variable bandwidth:
\begin{equation}\label{eq:kernel}
k_Q(x,x')=\exp\left({-\frac{| F_Q(x)-F_Q(x')|^2}{\delta}}\right)
\end{equation}
where $\delta$ is a positive bandwidth parameter that can vary based on the available data density in $\mathcal X$. We employ a class of variable bandwidth kernels, also known as self-tuning kernels, as introduced by \cite{Berry2016Variable}.
Associated with this square-integrable kernel $k_Q$ is a compact integral operator:
\begin{equation}\label{eq:integral_op}
K_Qf(x) := \int_\mathcal X k_Q(x,y)f(y)d\mu(y)
\end{equation}

The action of the kernel operator $K_Q$ is approximated through a matrix $K$ with the elements given as $K_{ij}=k_Q(x_i,x_j)$. The normalized Markov kernel $P$ is approximated through 
\begin{equation}\label{eq:MarkovKernelMatrix} P_{ij}=\frac{K_{ij}}{(\sum_{k=1}^{N}K_{ik}q_k^{-1/2})q_j^{1/2}}, \quad q_i=\sum_{k=1}^N K_{ik} \end{equation}
and it defines an orthonormal basis for the Koopman eigenfunctions, based on the data sampled along the orbit. We obtain an approximation of the eigenvalues $\lambda_j$ and eigenfunctions $\varphi_j$ of $P$ by solving the algebraic eigenvalue problem,
\begin{equation}\label{eq:KernelEigen} 
P\varphi_j=\lambda_j\varphi_j 
\end{equation}
A sparsification of $P$ can reduce the computational burden for large sample numbers $N$. This process involves selecting a cutoff value $k_{nn}\ll N$, setting all but the largest $k_{nn}$ elements in each row of $K$ to zero, and symmetrizing the resulting sparse matrix.

The Markov matrix $P$ has real eigenvalues ordered as $1=\lambda_1>\lambda_2\geq\lambda_3\geq...$, and real eigenvectors $\varphi_j$ that are mutually orthogonal in $\mathbb R^N$. Note that the first eigenvector corresponding to $\lambda_1=1$ is the constant eigenvector $\varphi_1=(1,...,1)^T\in\mathbb R^N$, which has zero Dirichlet energy. One key feature of $P$ is that as $Q\rightarrow\infty$, it commutes (and hence has common invariant subspaces) with the Koopman operator $U^t$. Therefore, Koopman eigenfunctions can be approximated by a linear combination of kernel operator eigenfunctions:
\begin{equation}\label{eq:KoopmanEigenfunction}
\psi_i=\sum_{j=1}^lc_{ji}\varphi_j, \quad i=1,\cdots,l 
\end{equation}
where $\psi_i\in\mathbb R^N$ denotes the $i$-th eigenfunction of $U^t$ sampled at $N$ data points. Hence, $l\leq N$ first eigenfunctions of the kernel operator are used to approximate the Koopman eigenfunctions.
The matrix $c_{l\times l}=[c_{ij}]$ is computed using a data-driven Galerkin approach \cite{Giannakis2019Galerkin}, leading to the following matrix generalized eigenvalue problem:
\begin{equation}\label{eq:GenEigen} A\hat c=\gamma B\hat c \end{equation}
where 
\begin{equation}\label{eq:GalerkinMatrices} 
\begin{aligned} A_{ij} &= \langle\varphi_i,V\phi_j\rangle-\epsilon\langle\varphi_i,\Delta\phi_j\rangle, \\
B_{ij} &= \langle\varphi_i,\phi_j\rangle, \quad \phi_i = \frac{\varphi_i}{\eta_i}, \\
\eta_i&=\frac{1}{\epsilon}(\lambda_i^{-1}-1) 
\end{aligned} \end{equation}
for $i,j=1,\cdots,l$. Here $\epsilon$ is a (small) regularization parameter, and each column of matrix $c$ corresponds to a solution of $\hat c$ in \eqref{eq:GenEigen}. Solutions $(\gamma_i,\psi_i)$ are placed in order of increasing Dirichlet energy, denoted by $E$, calculated for each eigenfunction $\psi_i$ as
\begin{equation}\label{eq:Dirich} 
E(\psi_i)=\frac{\langle\psi_i,\Delta\psi_i\rangle}{|\psi_i|^2} 
\end{equation}
The approximation in \eqref{eq:GalerkinMatrices} relaxes the exact eigenvalue assumption, allowing for the identification of modes that are coherent over finite time scales, where the system's dynamics exhibit less sensitivity to perturbations. This is analogous to the $\epsilon$-eigenvalue framework introduced by Colbrook et al. \cite{Colbrook2021}. Similarly, Dirichlet energy \eqref{eq:Dirich} serves as a proxy for selecting smooth and physically meaningful eigenfunctions that avoid spectral pollution.


After approximating the Koopman eigenfunctions at the sampled data points $X_s={x_1,\cdots,x_N}$, we utilize the Nyström extension approach to extend the Koopman eigenfunctions to points not included in the original dataset, i.e. $x\in \mathcal X\setminus X_s$. To achieve this, we first apply the Nyström approach to extend the eigenfunctions $\varphi_k$ of the Markov kernel operator. Then, we use equation \eqref{eq:KoopmanEigenfunction} for out-of-sample evaluation of Koopman eigenfunctions. Our approach is based on the method established in \cite{Coifman2006Nystrom}.

\subsection{Forecasting by Koopman eigenfunctions}

The linear property of Koopman operator allows us to construct a linear evolution model for Koopman eigenfunctions:
\begin{equation}\label{eq:KoopmanEvolution}
\psi(x(t+1))=K_{\psi}\psi(x(t)) 
\end{equation}
where $\psi=(\psi_1,...,\psi_{l'})^T$ is a vector of Koopman eigenfunctions and $K_{\psi}$ is an $l'\times l'$ matrix computed from the data. We construct two matrices using the approach described in Section \ref{sec:data-driven}: $\Psi^-=[\psi(x_1),...,\psi(x_{N-1})]$ and $\Psi^+=[\psi(x_2),...,\psi(x_N)]$. The matrix $K_{\psi}$ is then determined by solving:
\begin{equation}\label{eq:OptimizationKpsi} 
\min_{K_\psi} |\Psi^+ - K_\psi\Psi^-|_2 
\end{equation}
which is solved using normal equations.

Next, we propose a linear decoder that maps Koopman eigenfunctions to system states. Assuming a sufficiently large set of Koopman eigenfunctions spans the system state space, we express the state as:
\begin{equation}\label{eq:StateReconstruction} 
x = C\psi(x) 
\end{equation}
where $x \in \mathbb{R}^d$ is the system state vector and $C$ is a $d\times l'$ matrix. We determine $C$ by solving:
\begin{equation}\label{eq:DecoderOptimization} \min_C |X - C\Psi|_2 \end{equation}
where $X=[x_1,...,x_N]$ and $\Psi=[\psi(x_1),...,\psi(x_N)]$ are time-ordered matrices.

\section{Clustering the data} \label{sec:cluster}
We employ a clustering-based Koopman approach to group power stations based on their temporal power patterns and then predict future energy production for each pattern. Generally, power generation follows a consistent pattern of evolution over time, which can be identified by mapping power temporal data to a lower-dimensional space.
Dimensionality-reduction methods such as PCA \cite{Moon2000Book}, diffusion maps \cite{Coifman2006Diffusion}, and t-SNE \cite{t-NSE2008} are popular tools for the visualization of high-dimensional data. However, these methods have some limitations, such as sensitivity to noise, scrambled global structure of data, computational scalability, and optimization for two- or three-dimensional visualization. To address these concerns, a method based on the potential of heat diffusion for affinity-based transition embedding called PHATE was recently developed \cite{PHATE2019}. While the original work focused on biological data, in this study, we apply PHATE to detect distinct power generation patterns.
We consider a min-max normalization $\hat x$ of a typical time-ordered vector $x$.
For clustering, the load data at $d$ grid points are time-ordered vectors in $\mathbb R^N$. We denote the load recorded at station $i$ as $y_i=[y_i(t_1),\cdots,y_i(t_N)]^T$, where $i=1,...,d$ and measure the empirical similarities between different load patterns for the normalized vectors $\hat y_i$ using a variable bandwidth kernel. This leads to a kernel matrix, denoted as $\hat P_{d\times d}$, where the elements $\hat P_{ij}$ indicate the similarity between each pair $(\hat y_i,\hat y_j)$.
It is worth noting that evaluating similarities of spatial points does not involve any time-delay $Q$. Also, the matrix $\hat P$ used in PHATE to compute similarities between different time points differs from the matrix $P$ used in Section \ref{sec:data-driven}--due to the difference between the vectors $y_i\in\mathbb R^N$ and $x_i\in\mathbb R^d$.
After defining an initial diffusion operator like $\hat P$, PHATE proceeds with a few carefully designed steps. Firstly, the diffusion operator is raised to its $t'$-th power, where the optimal diffusion time scale $t'$ is chosen based on von Neumann entropy. This provides a robust intrinsic data distance that preserves the global structure of the data and low-pass filters the data. To embed the global structure in low (two or three) dimensions and resolve instabilities, PHATE considers $\hat U^{t'}_i=-\log(\hat P^{t'}_i)$, where $\hat P^{t'}_i$ is the $i$-th row of $\hat P^{t'}$, and computes the potential distances:
\begin{equation}\label{eq:potential}
\Gamma^{t'}_{ij}=|\hat U^{t'}_i-\hat U^{t'}_j|_2
\end{equation}
The final step involves performing a metric multidimensional scaling (metric MDS) to embed the data into a low-dimensional space (in our case, a 3-dimensional space). This is done by minimizing the following 'stress' function:
\begin{equation}\label{eq:stress}
\text{stress}(\hat z_1,...,\hat z_d)=\sqrt{\frac{\sum_{i,j}(\Gamma^{t'}{ij}-|\hat z_i-\hat z_j|)^2}{\sum{i,j}(\Gamma^{t'}_{ij})^2}}
\end{equation}
where the function operates on the embedded $m$-dimensional coordinates of data points. While the classical MDS assumes that input distances correspond to low-dimensional Euclidean distances, this assumption is overly restrictive for the PHATE setting. The metric MDS, however, relaxes this assumption by requiring only that the input distances be a distance metric. 


\section{Results}\label{sec:Results}
\begin{figure}[!]
\centering
\subfloat[\label{fig:Eigenfunction}]{\includegraphics[clip,width=1\columnwidth]{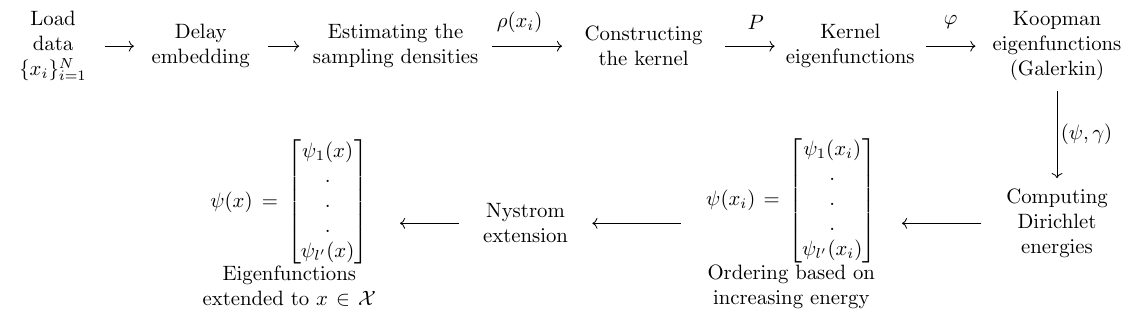}} \
\subfloat[\label{fig:Pred}]{\includegraphics[clip,width=1\columnwidth]{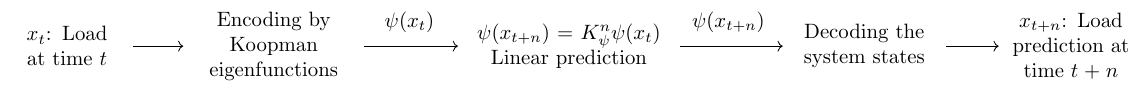}}
\caption{(a) Computing Koopman eigenfunctions. (b) Load forecasting using Koopman.}
\label{fig:overview}
\end{figure}

Here, we present results from applying the Koopman operator approach to the data described in Section \ref{sec:data}. Figure \ref{fig:overview} illustrates a schematic of our proposed numerical approach for load forecasting using Koopman eigenfunctions. The performance of our approach is evaluated using the root mean squared error (RMSE).
\begin{equation}\label{eq:RMSE}
\text{RMSE}=\sqrt{\frac{1}{n_x}\sum_{i=1}^{n_x}(x_i-x^f_i)^2}
\end{equation}
which applies to a vector $x=[x_1 \ ... \ x_{n_x}]^T$ and its forecasted value $x^f=[x^f_1 \ ... \ x^f_{n_x}]^T$. Throughout this paper, all error and RMSE values were computed using min-max normalized load vectors, limiting the RMSE to the range of 0 to 1.
\subsection{Data description}\label{sec:data}
\begin{figure}[!]
\centering
\subfloat[\label{fig:map}]{\includegraphics[clip,width=0.2\columnwidth]{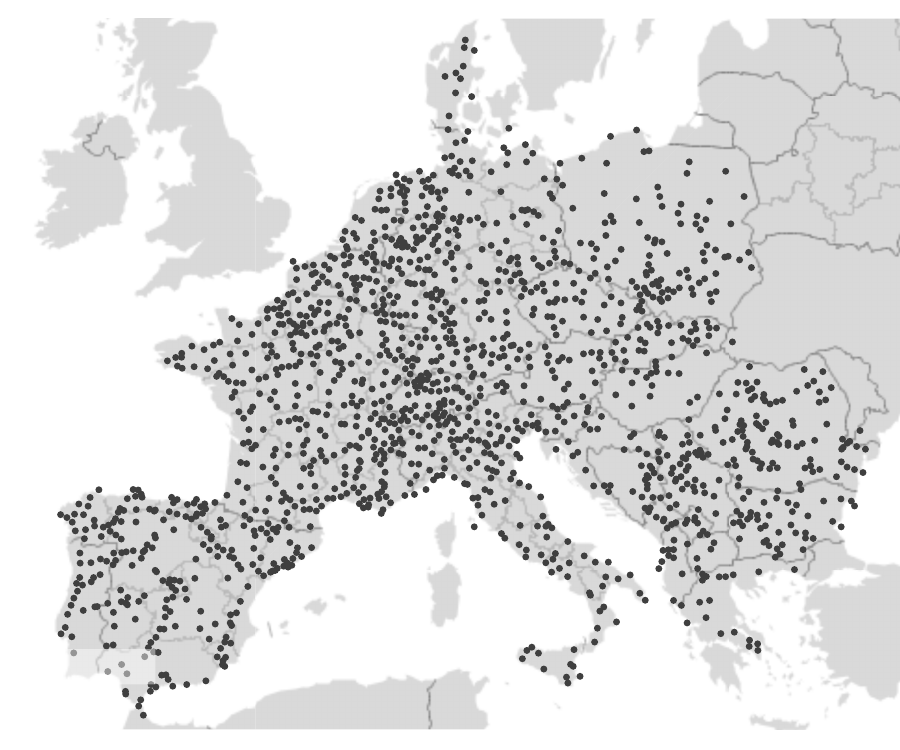}} \quad
\subfloat[\label{fig:net}]{\includegraphics[clip,width=0.2\columnwidth]{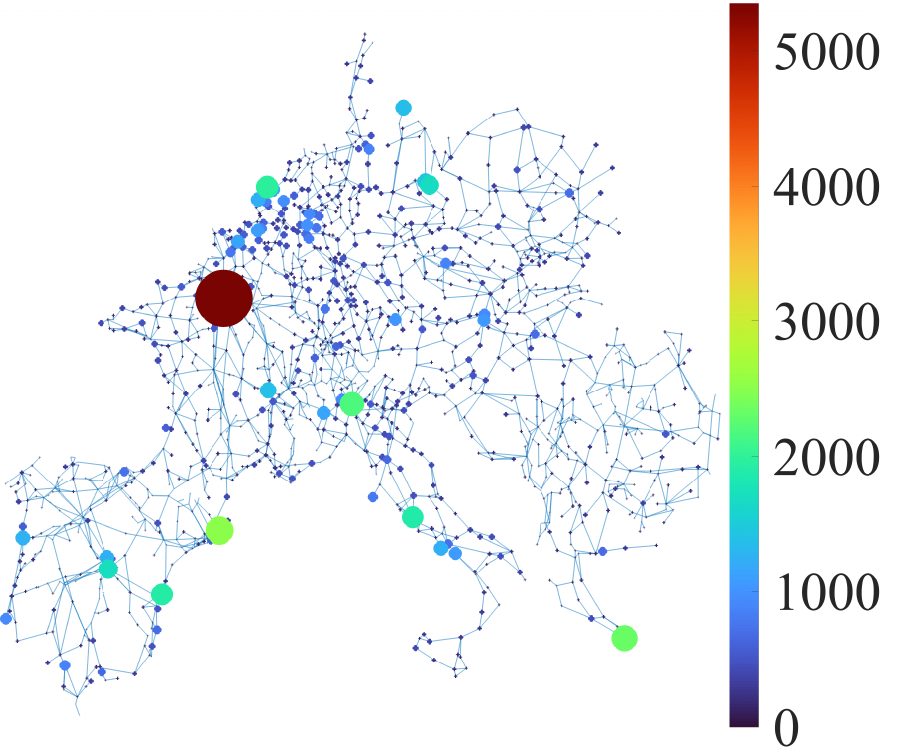}} \quad
\subfloat[\label{fig:MeanLoad}]{\includegraphics[clip,width=0.5\columnwidth]{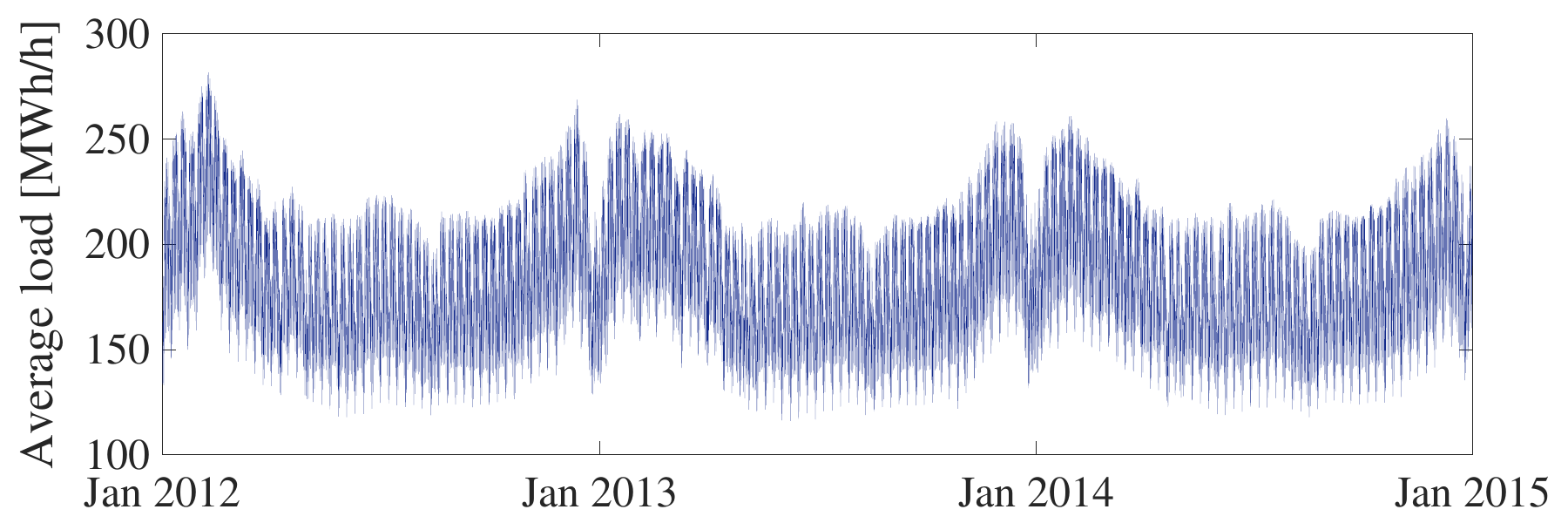}}
\caption{Overview of dataset. (a) Distribution of generators. (b) Transmission grid and the 3-year average power of each generator in MWh/h. (c) Average load of all power plants. The sampling rate is 1 hour.}
\label{fig:DataOverview}
\end{figure}

This paper utilizes a dedicated and large-scale data set for a renewable electric power system \cite{Jensen2017Data} from the continental European electricity system collected for the period of 2012-2014 and includes data on demand and renewable energy inflows with a high resolution of 50 km and 1 hour.
The dataset covers mainland Europe over 3 years and consists of a transmission network model, as well as information for generation and demand. The data was collected from 1494 transmission buses with 2156 lines, including conventional generators with their technical and economic characteristics, as well as weather-driven forecasts and corresponding realizations for renewable energy generation. Figure \ref{fig:DataOverview} provides an overview of the dataset and the power grid.

\subsection{Reduced-order modeling of the load dynamics}
\begin{figure}[!]
\centering
\includegraphics[clip,width=1\columnwidth]{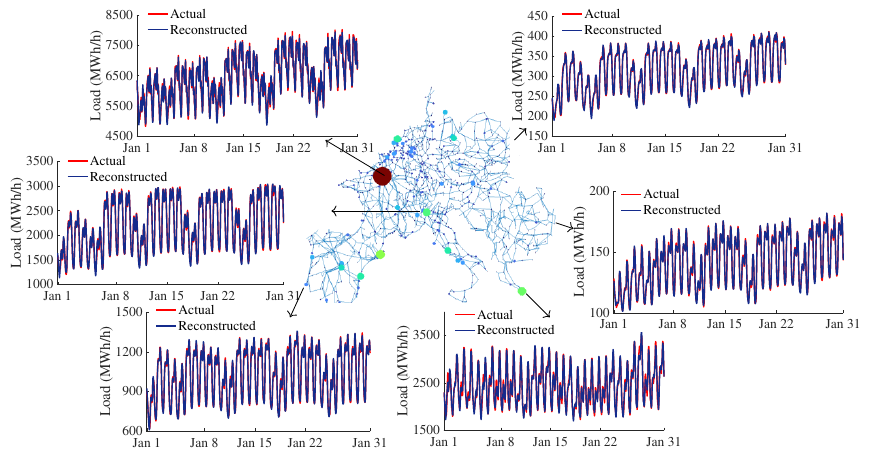}
\caption{Examples of spatiotemporal load reconstruction. The load data of 1494 stations during January 2014 are reconstructed using 100 Koopman eigenfunctions.}
\label{fig:ReconJan2014}
\end{figure}

By applying our approach to load data from January 2014 with a delay of $Q=500$ and a regularizing parameter of $\epsilon=10^{-9}$, we identified 100 Koopman eigenfunctions and used them to reconstruct the load for all 1494 stations during that month. Example load reconstructions are shown in Figure \ref{fig:ReconJan2014}. This same approach can be used to reconstruct the load for other months or over larger time horizons, though doing so may require increasing the length of the delay, which presents computational challenges due to the large amounts of data involved.
\subsection{Nonlinear spatiotemporal spectral analysis of the load dynamics}
\begin{figure}[!]
	\centering
	\subfloat{\includegraphics[clip,width=0.3\columnwidth]{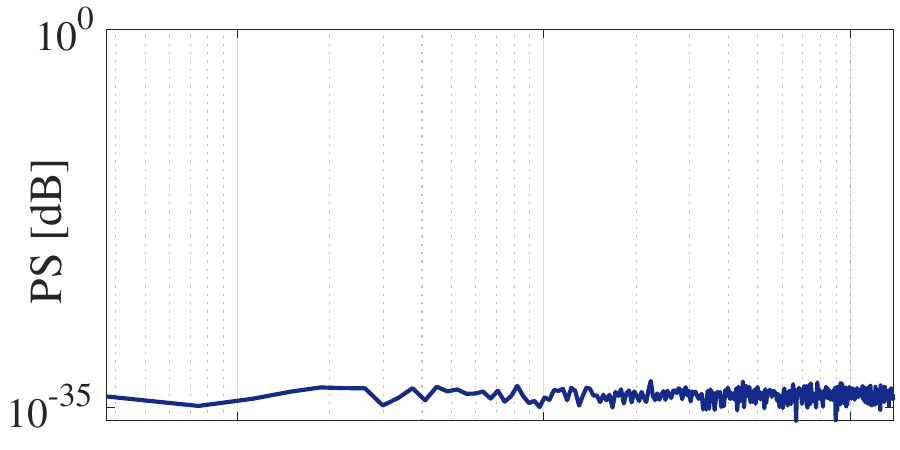}} \ \
	\subfloat{\includegraphics[clip,width=0.3\columnwidth]{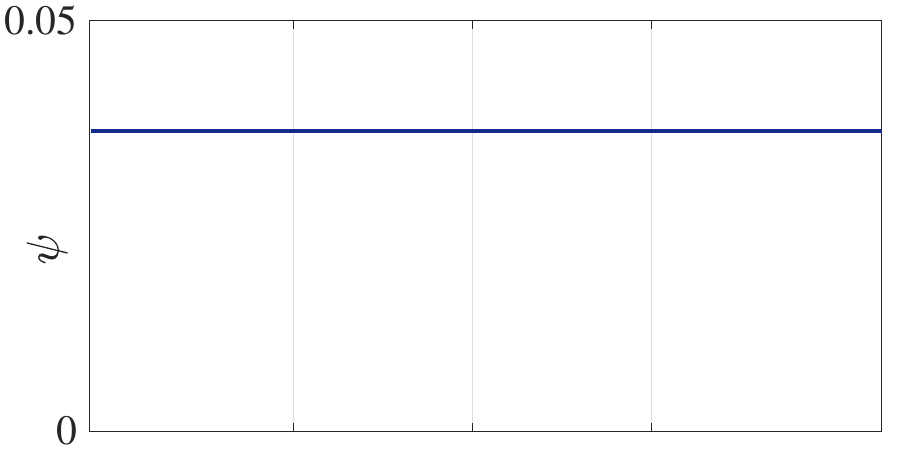}}
	\subfloat{\includegraphics[clip,width=0.2\columnwidth]{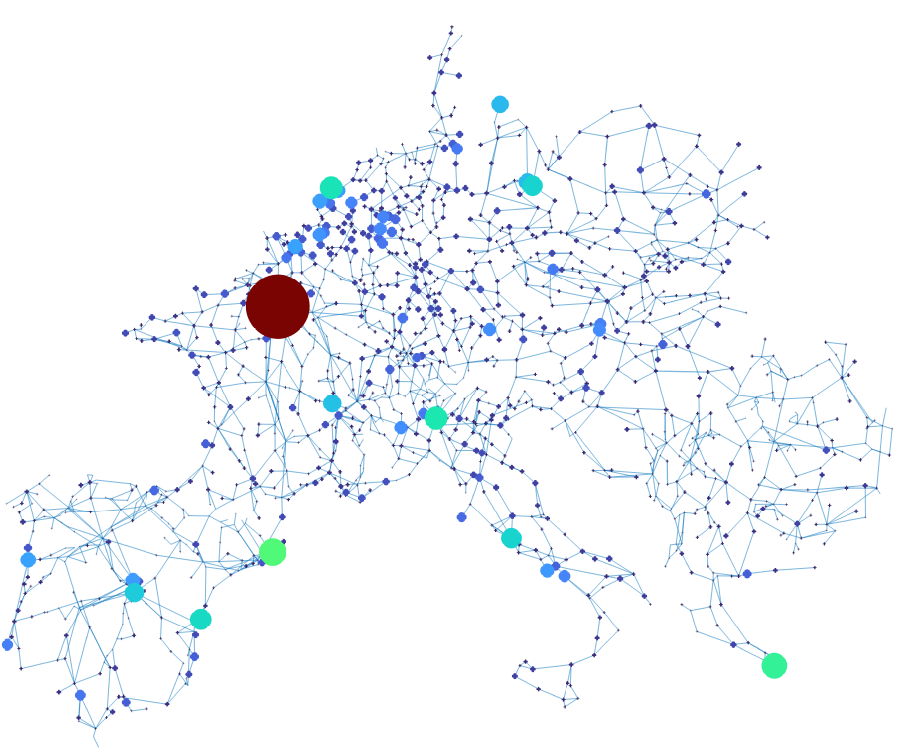}} \\
	\subfloat{\includegraphics[clip,width=0.3\columnwidth]{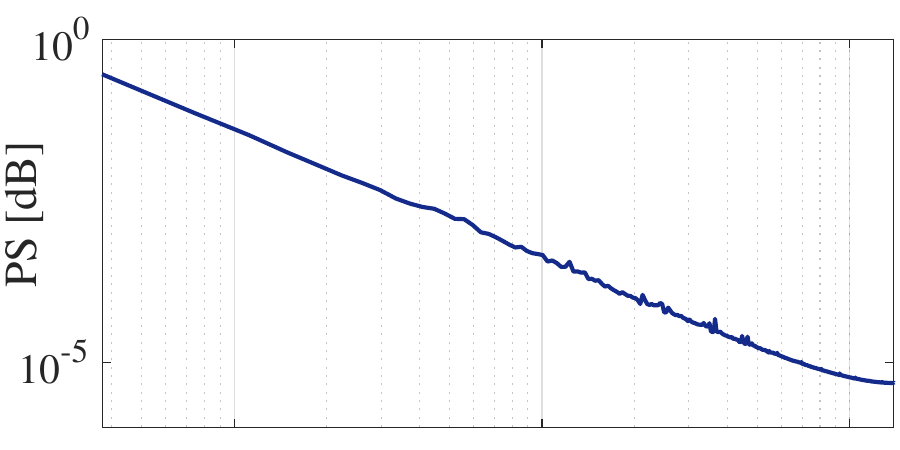}} \ \
	\subfloat{\includegraphics[clip,width=0.3\columnwidth]{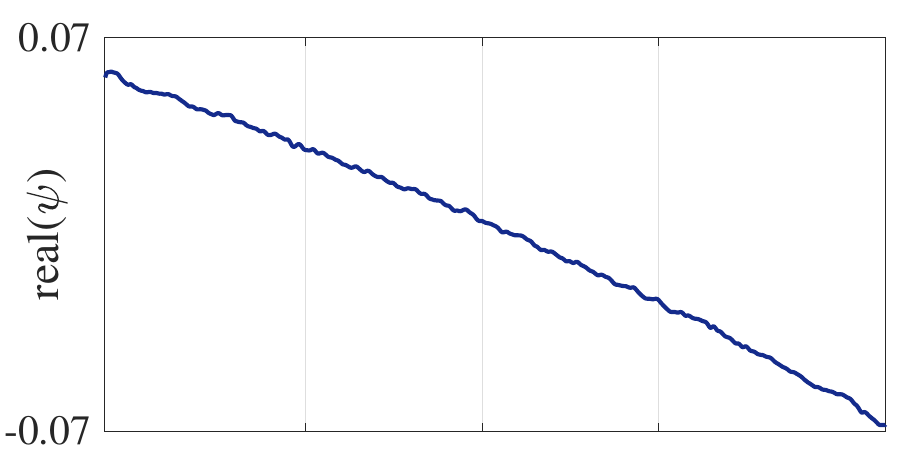}}
	\subfloat{\includegraphics[clip,width=0.2\columnwidth]{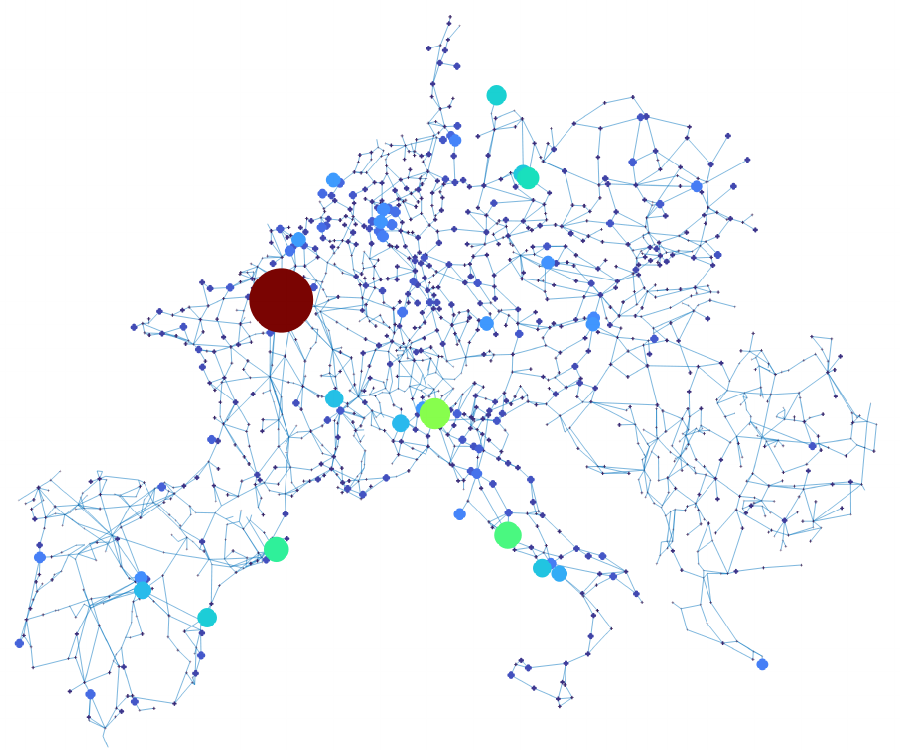}} \\
	\subfloat{\includegraphics[clip,width=0.3\columnwidth]{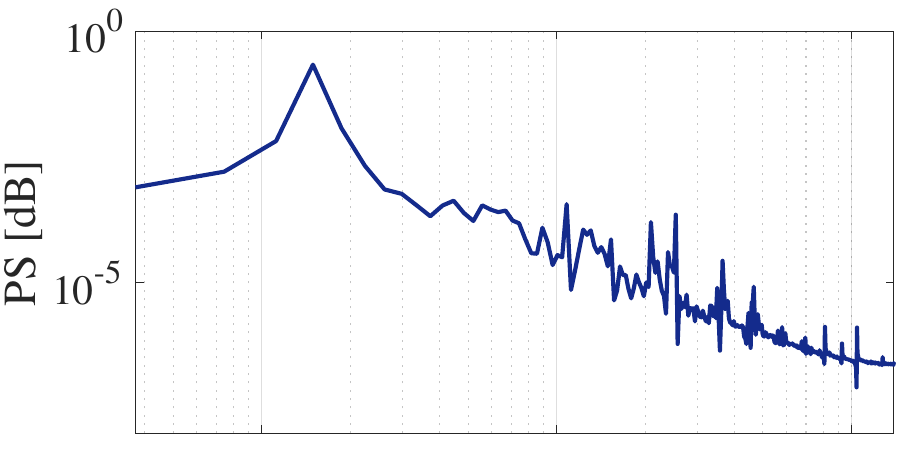}} \ \
	\subfloat{\includegraphics[clip,width=0.3\columnwidth]{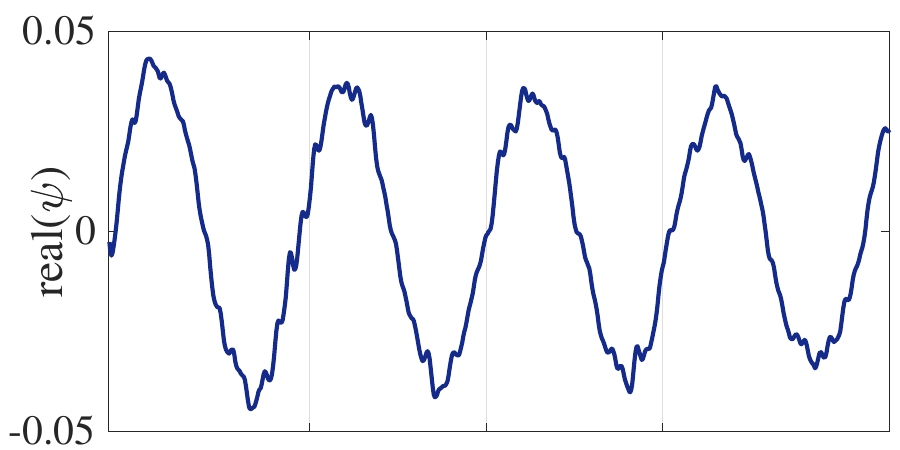}}
	\subfloat{\includegraphics[clip,width=0.2\columnwidth]{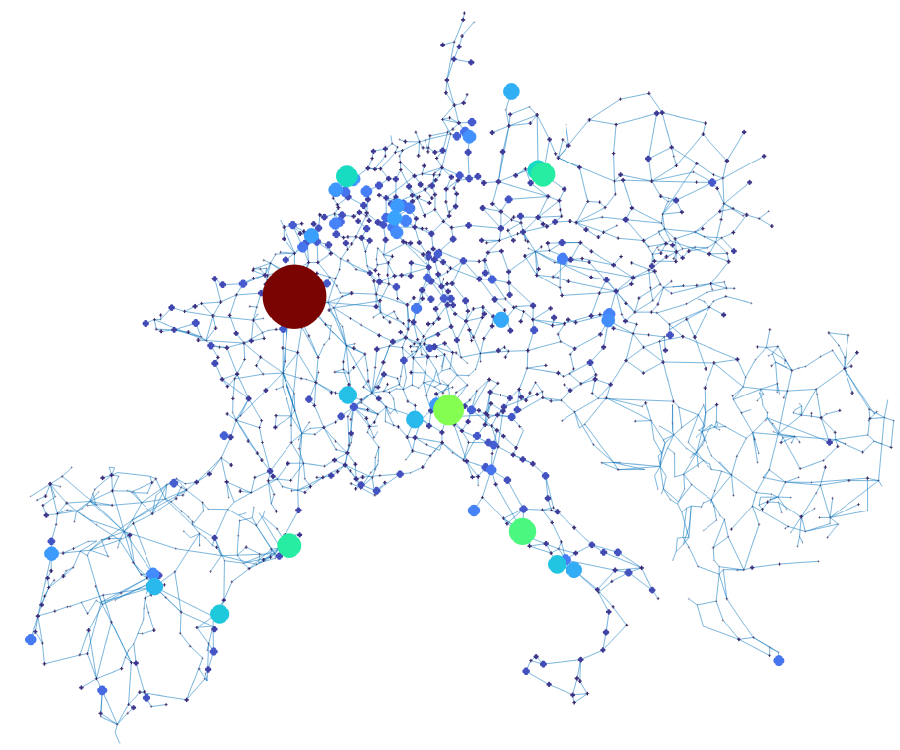}} \\
	\subfloat{\includegraphics[clip,width=0.3\columnwidth]{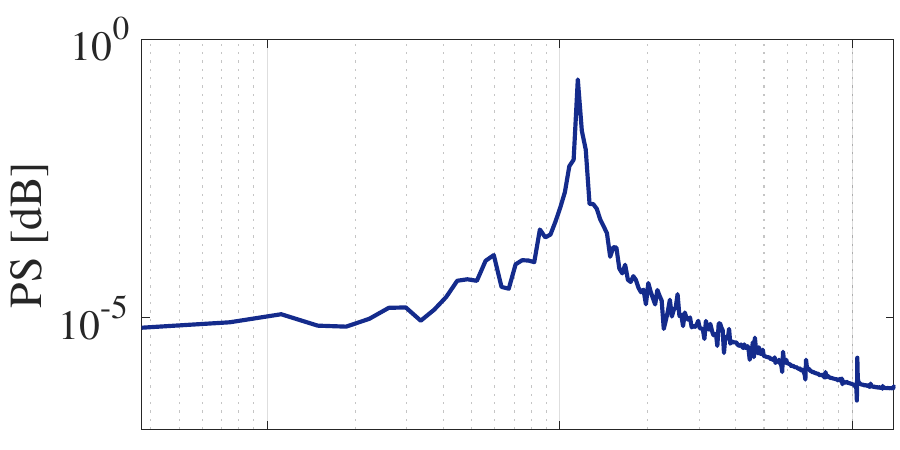}} \ \
	\subfloat{\includegraphics[clip,width=0.3\columnwidth]{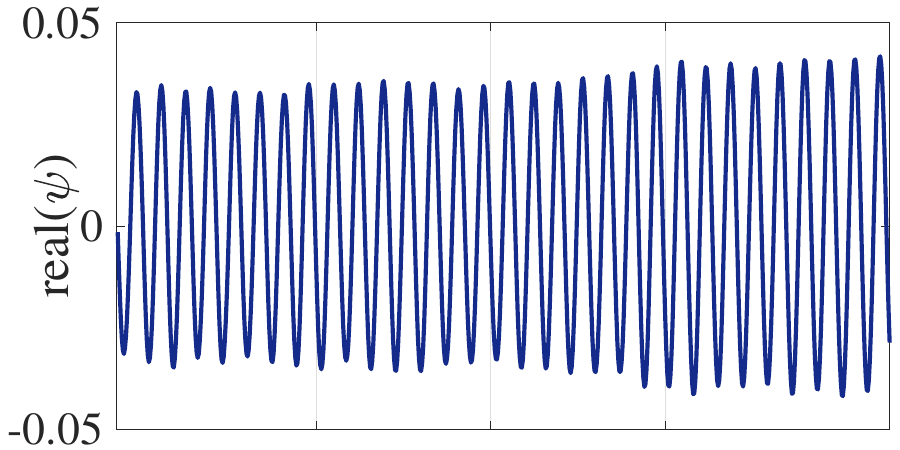}}
	\subfloat{\includegraphics[clip,width=0.2\columnwidth]{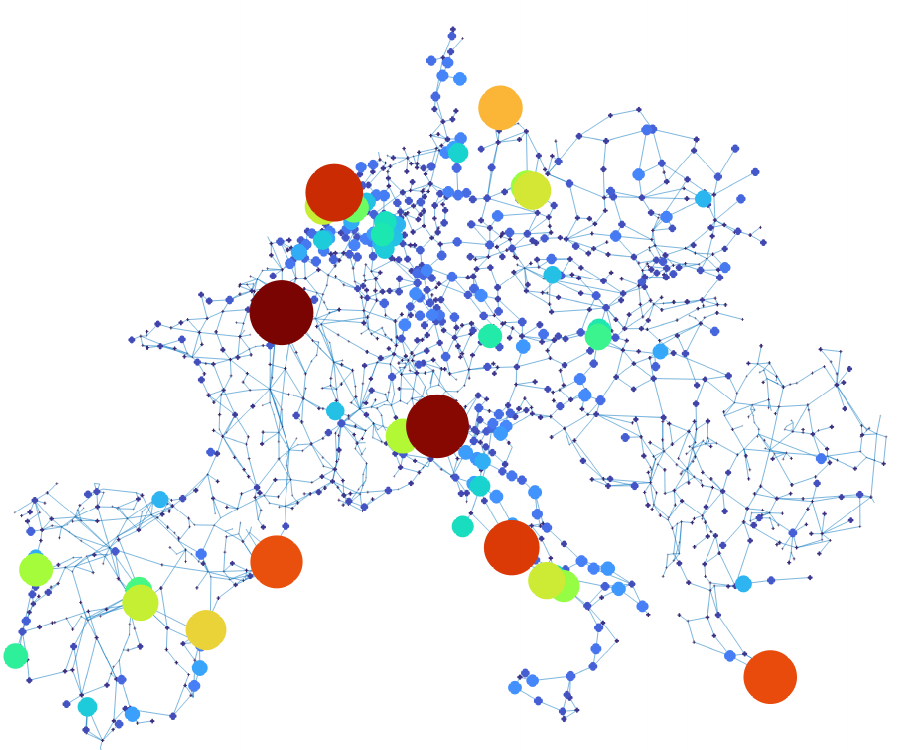}} \\
	\subfloat{\includegraphics[clip,width=0.333\columnwidth]{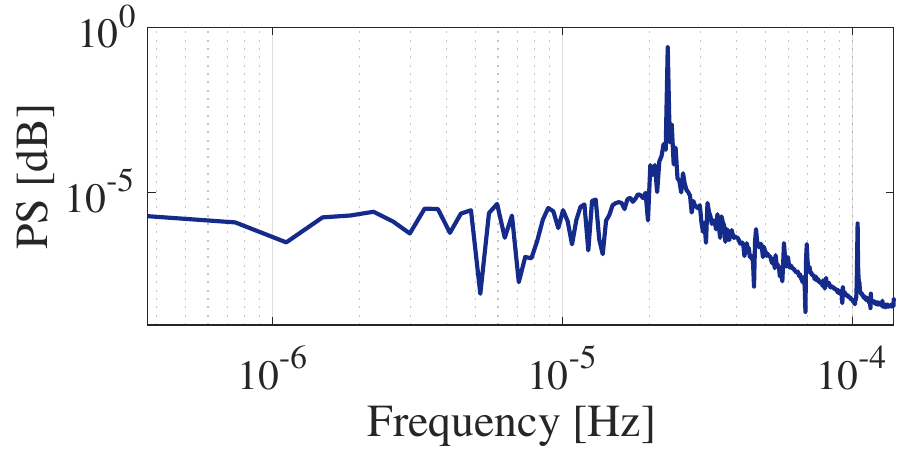}} \ \
	\subfloat{\includegraphics[clip,width=0.333\columnwidth]{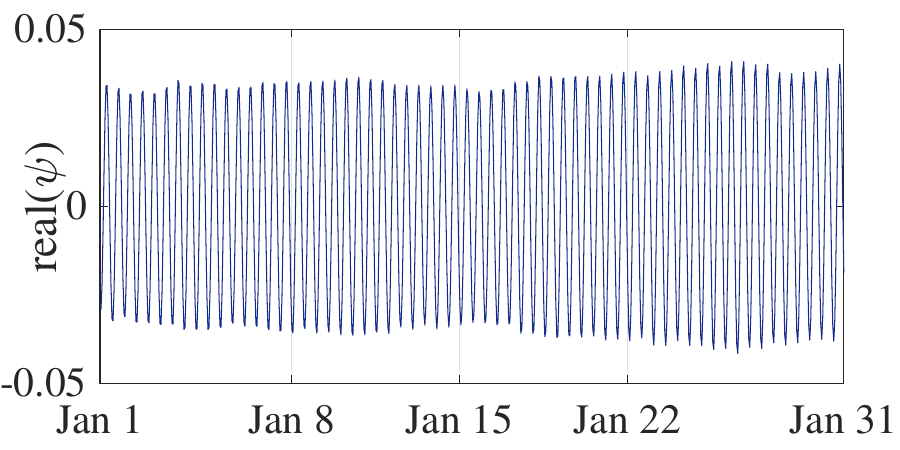}}
	\subfloat{\includegraphics[clip,width=0.2\columnwidth]{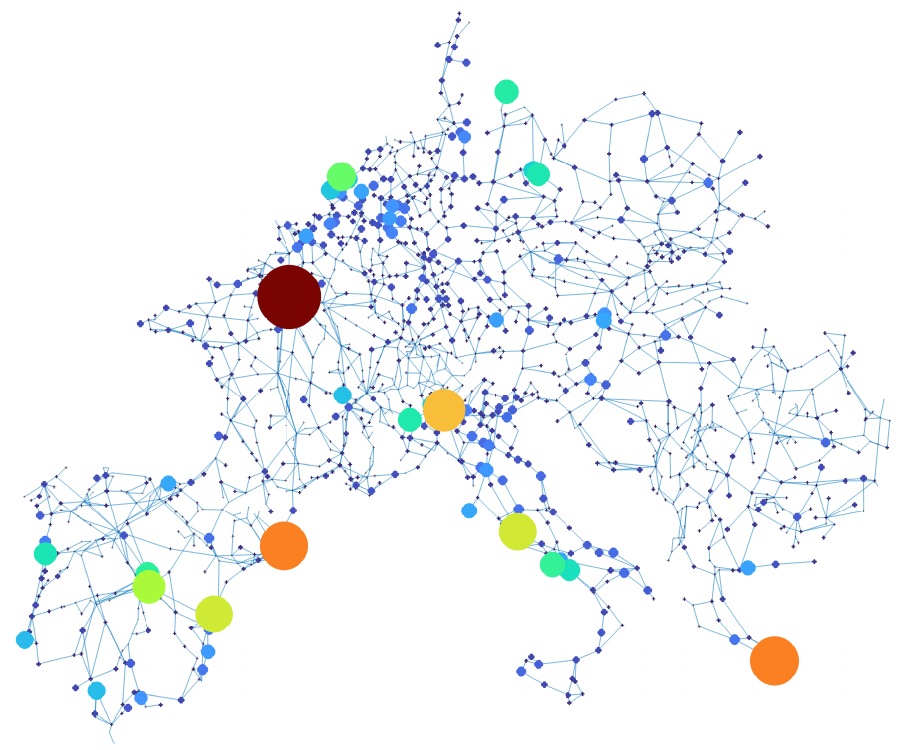}} 
	\caption{Spatial and temporal patterns with multiple scales recovered by the load data during January 2014. The left, middle, and right columns display the Power Spectrum (PS), time evolution, and spatial pattern induced by the Koopman eigenfunction, respectively. The first row represents the trivial Koopman eigenfunction ($\omega=0$), followed by subsequent rows associated with frequencies scaled with a month ($3.86\times 10^{-7} \text{Hz}$), a week ($1.65\times 10^{-6} \text{Hz}$), a day ($1.15\times 10^{-5} \text{Hz}$), and a half day ($2.31\times 10^{-5} \text{Hz}$), arranged in ascending order.}
	\label{fig:ModesJan2014} 
\end{figure}

We apply the algorithm to obtain coherent patterns associated with each month. For example, Figure \ref{fig:ModesJan2014} shows these patterns, highlighting the Koopman eigenfunctions with the lowest Dirichlet energy values for January 2014. While temporal modes of different months are similar, the spatial modes associated with different months might be generally different (see Figure \ref{fig:SpatialPatternDay}).
Figure \ref{fig:ModesJan2014} reveals that the dominant frequencies contributing to load dynamics scale with zero, one month, one week, one day, and half a day. These load dynamic components have physical interpretations. The zero-frequency mode corresponds to the average loads generated in individual power stations (compare the spatial pattern in the first row of Figure \ref{fig:ModesJan2014} to average loads in Figure \ref{fig:DataOverview}). The mode scaling with one month in the second row of Figure \ref{fig:ModesJan2014} is due to monthly weather changes, e.g., temperature and solar radiation. Additionally, the modes in the third row of Figure \ref{fig:ModesJan2014}, representing a weekly frequency, are influenced by working days and holidays, while the mode scaling in the fourth row of Figure \ref{fig:ModesJan2014}, corresponding to a daily frequency, is affected by changes in solar radiation over 24 hours. The final row in Figure \ref{fig:ModesJan2014} displays a half-day frequency, which indicates variations in power usage patterns during day and night.


The middle column of Figure \ref{fig:ModesJan2014} illustrates that the nonlinear load signal can be accurately reconstructed through a small number of highly predictable modes that exhibit near-periodic behavior. In contrast, the left column of Figure \ref{fig:ModesJan2014} shows that each mode has a narrow frequency band, reflecting intermittent patterns that play a crucial role in capturing features of strongly nonlinear dynamics \cite{Giannakis2012NLSA}. 

\subsection{Power usage patterns detected by spatial modes}
The right column in Figure \ref{fig:ModesJan2014} reveals that each Koopman eigenfunction elicits a distinct spatial pattern for its corresponding mode. While the spatial modes linked to zero and low frequencies in the upper three rows of Figure \ref{fig:ModesJan2014} are dominated by the station that generates the largest average value in Figure \ref{fig:DataOverview} (a station in France), the two last rows highlight that high-frequency modes, scaled with a day and a half day, produce dissimilar patterns.
Specifically, the mode scaled with a day shows significant amplitudes in two stations in Northern Europe (Germany) and additional stations in Southern Europe (Italy, Greece, and Spain). Moreover, as observed from the last row of Figure \ref{fig:ModesJan2014}, these southern stations, second only to the one in France, generate the highest values in the mode scaled with a half day.
This pattern suggests that different stations generate power for different purposes: stations producing power for industrial usage represent the largest values in low-frequency modes, while stations supplying power for residential and commercial areas manifest in the higher-frequency (day and half day) modes.

\begin{figure}[!]
	\centering
	\subfloat{\includegraphics[clip,width=0.25\columnwidth]{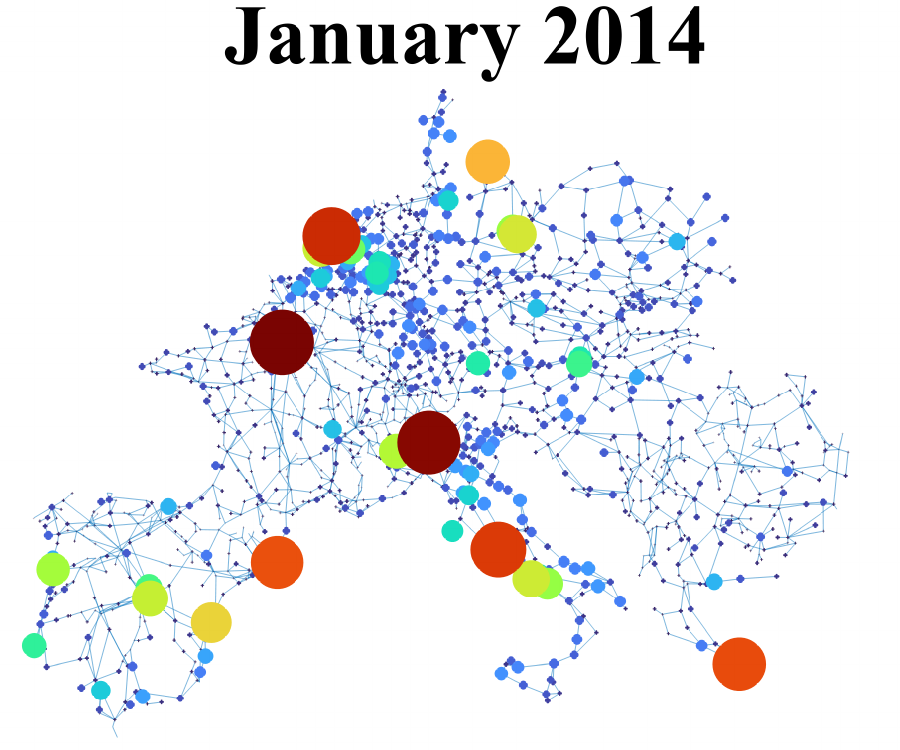}} 
	\subfloat{\includegraphics[clip,width=0.25\columnwidth]{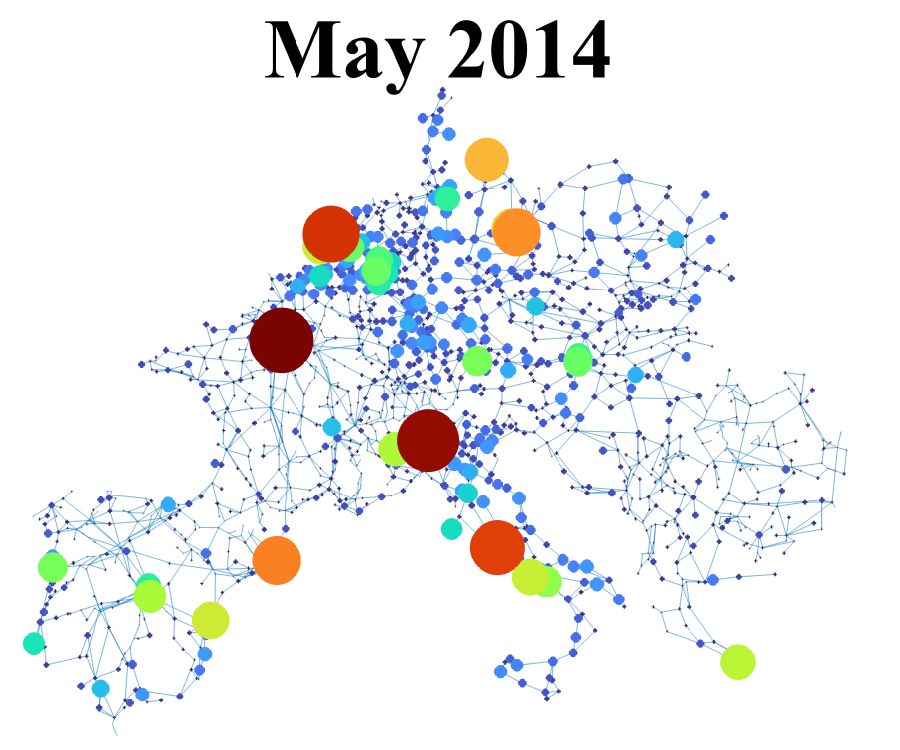}} 
	\subfloat{\includegraphics[clip,width=0.25\columnwidth]{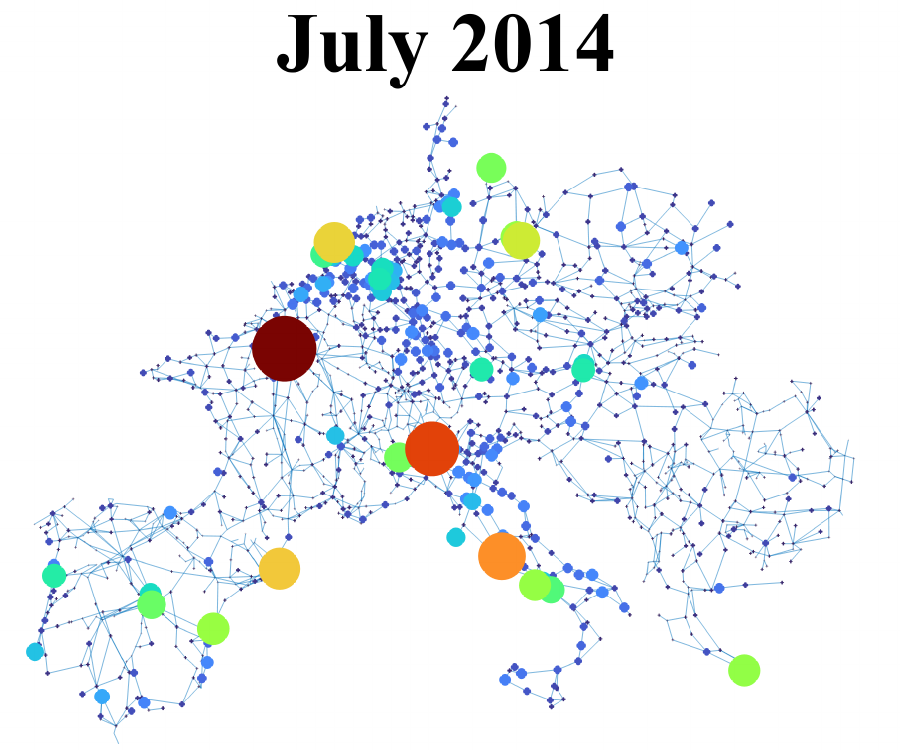}} 
	\subfloat{\includegraphics[clip,width=0.25\columnwidth]{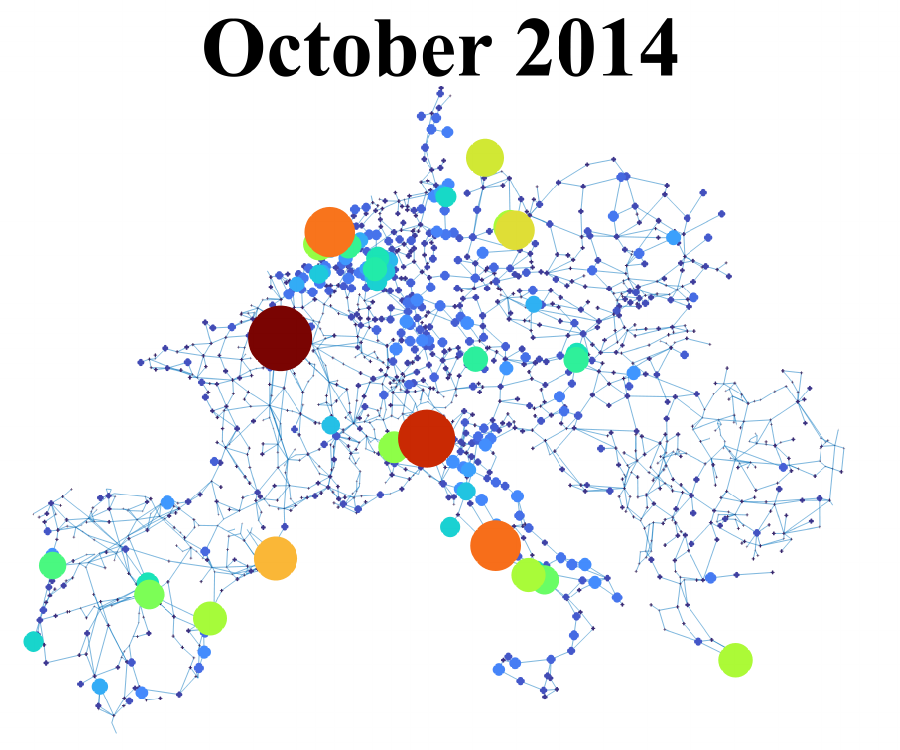}} 
	\caption{Example spatial patterns showing the frequency scaled with a day.}
	\label{fig:SpatialPatternDay} 
\end{figure}

Examining spatial modes across various months reveals seasonal variations in energy patterns. Figure \ref{fig:SpatialPatternDay} displays the spatial patterns corresponding to frequencies scaled over a 24-hour period for January, May, July, and October of 2014, representing winter, spring, summer, and fall, respectively. Notably, certain stations in Southern Europe exhibit distinctive modes during the winter months, but their contribution diminishes during the warmer months from June to October 2014. Consequently, power usage in these regions remains relatively constant throughout the day in warmer months but varies significantly during colder months.


\subsection{Clustering the data}\label{sec:dataClust}

\begin{figure}[!]
	\centering
	\subfloat{\includegraphics[clip,width=0.5\columnwidth]{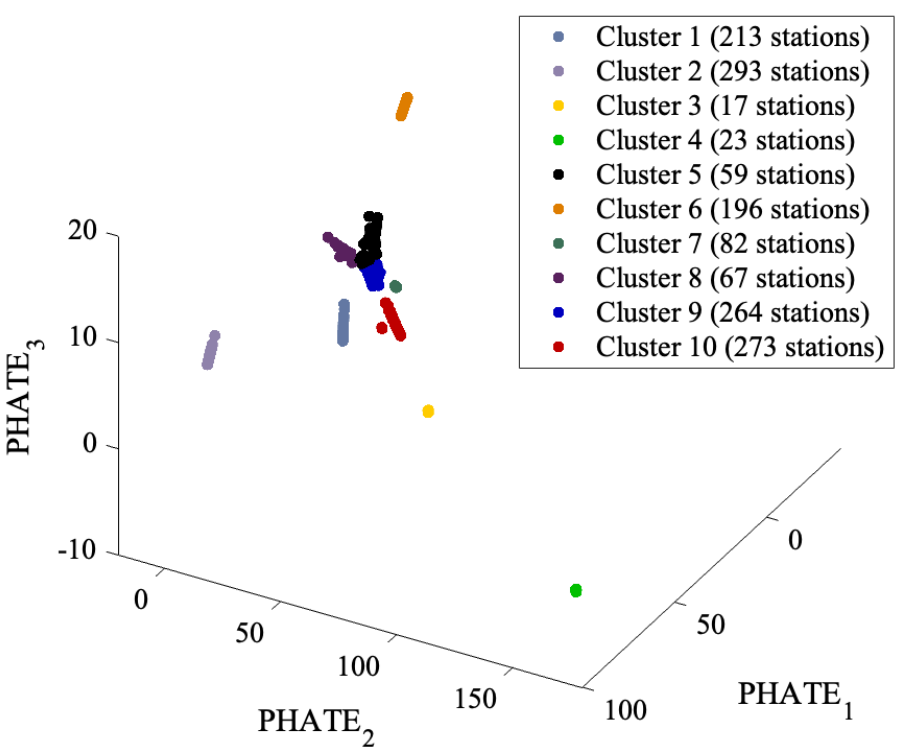}} \\
	\subfloat{\includegraphics[clip,width=1\columnwidth]{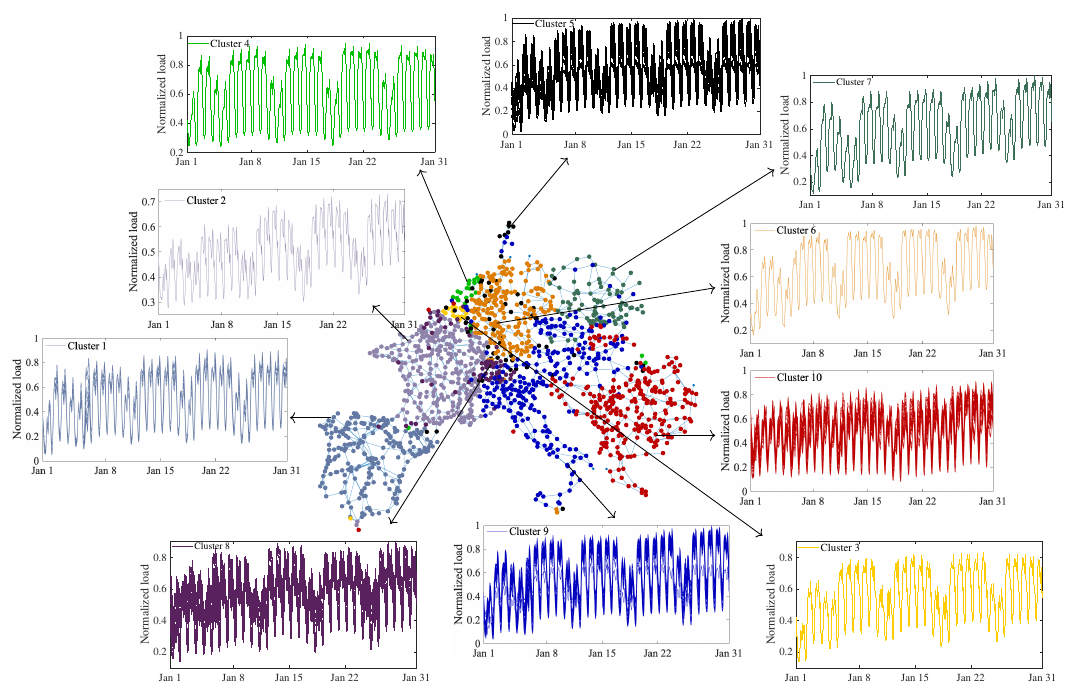}} 
	\caption{Visualizing the data on the low-dimensional PHATE coordinates unveils 10 clusters.}
	\label{fig:Cluster} 
\end{figure}

Figure \ref{fig:Cluster} shows the results of applying the algorithm to normalized data from 2012-2014, where the upper plot displays the 10 clusters identified using the $k$-means algorithm \cite{Bishop2006}.
The lower plot in Figure \ref{fig:Cluster} demonstrates a clustering pattern that correlates with the physical distribution of power plants, with stations in different geographical areas falling into distinct clusters.
Figure \ref{fig:Cluster} also reveals clusters with completely synchronized stations, specifically clusters 2, 3, 4, 6, and 7. These correspond to groups that distinctly separate from the main body in the upper plot of Figure \ref{fig:Cluster}.
Additionally, clusters 1 and 10 are separated from the main body in the upper plot of Figure \ref{fig:Cluster}. This separation allows for further refinement of synchronized station groups, either by increasing the number of clusters in the $k$-means approach or by re-clustering the current clusters.
While increasing the number of clusters could potentially improve accuracy, we opt for 10 clusters as a trade-off between accuracy and model complexity. This clustering approach enables the identification of different power generation patterns and effectively divides the large-scale system into several interconnected subsystems.

\subsection{Short-term load forecasting}\label{sec:short}
Short-term load forecasting plays a crucial role in daily and weekly dispatching of power grids \cite{Kong2019LSTM}. To forecast a week ahead, we utilize Koopman modes derived from the previous week's load data. This approach assumes that no significant changes in Koopman modes occur between two successive weeks. However, substantial changes due to factors such as weather fluctuations or holidays may introduce errors. Despite this limitation, our proposed approach outperforms deep learning-based load forecasting methods.

To facilitate forecasting, we employ a hierarchical approach to cluster power stations. Load predictions are then generated for individual stations within each cluster through the construction of a single Koopman model for that cluster. As detailed in Section \ref{sec:dataClust}, we construct 10 Koopman models based on our analysis of load data, balancing model complexity with forecasting accuracy.

The numerical evaluation of the Koopman operator requires tuning several parameters, with the number of delays $Q$ and Koopman eigenfunctions $l'$ being particularly influential. Increasing the number of delays improves the extraction of highly coherent dynamical behaviors, and in the limit of infinitely many delays, one can recover the exact dynamical system properties \cite{Giannakis2019Delay}.
However, practical limitations in data availability restrict the number of delays that can be used. Furthermore, for finite-dimensional Koopman subspaces, the delay embedding window shows some correlation with the frequency ranges of interest \cite{Giannakis2019Delay}. Consequently, including uncorrelated delays may lead to overlooking important dynamical features within these ranges.

\begin{figure}[!]
	\centering
	\subfloat[\label{fig:RMSE_Q}]{\includegraphics[clip,width=0.45\columnwidth]{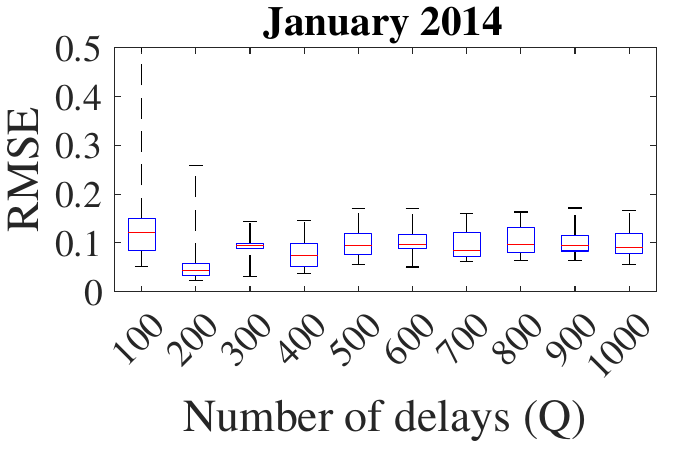}} \hspace{20pt}
 \subfloat[ \label{fig:RMSE_l}]{\includegraphics[clip,width=0.45\columnwidth]{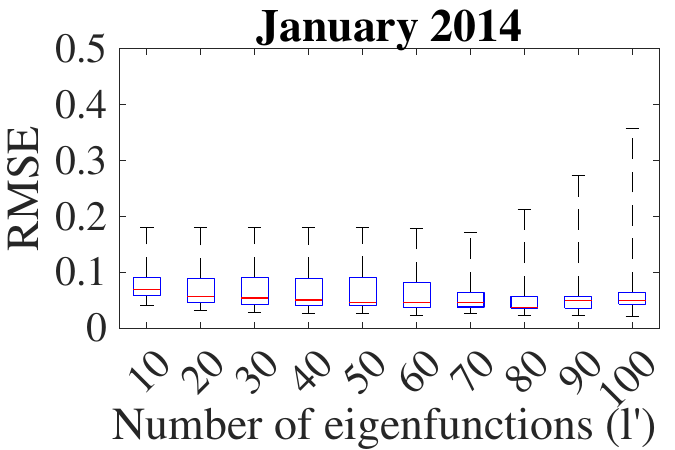}}
	\caption{RMSE of different stations at varying values of (a) time delay ($l'=100$), and (b) number of Koopman eigenfunctions for $Q=168 hr$ (one-week delay).}
	\label{fig:RMSE_Ql} 
\end{figure}

To investigate the role of delay horizon $Q$ in load forecasting, we fix the number of eigenfunctions and compute the RMSE for each station. Figure \ref{fig:RMSE_Q} shows the results for January 2014 as an example. This figure exemplifies the typical behavior observed across all months, where the RMSE values reach their minimum at around $Q = 200$. Notably, the RMSE values remain relatively stable as $Q$ increases beyond 200, highlighting the robustness of the proposed technique with respect to this pivotal parameter.
Given a sampling rate of one hour, 168 samples are available per week. This suggests that a delay time proportional to the forecasting horizon (one week in this instance) is reasonable for short-term forecasting. Therefore, we recommend setting the delay to scale with the forecasting horizon.
 
The number of Koopman eigenfunctions included in the forecasting model is the next parameter we investigate. While a rich library of eigenfunctions theoretically captures dynamical features across a wide spectrum, this does not necessarily translate to arbitrarily large libraries for real-world data. Including Koopman eigenfunctions with higher energy levels is often associated with computational errors and instability, potentially leading to false information.
In practice, these higher modes may not be accurately approximated from finite datasets, as they are not well-behaved on the data manifold \cite{Giannakis2019Spectral}. Additionally, the use of modes with little variance and no physical significance may not improve model interpretability. Overfitting in the regression problem \eqref{eq:OptimizationKpsi} is also a concern when the library $\Psi$ is excessively large \cite{Giannakis2019Spectral,Tavasoli2023koopman}.

Figure \ref{fig:RMSE_l} presents the RMSE values for varying values of $l'$ with a delay horizon of $Q=168$. The example from January 2014, shown in Figure \ref{fig:RMSE_l}, illustrates the typical behavior observed across all months: the RMSE reaches its minimum (both in terms of the RMSE distribution of individual stations and the mean value) at intermediate $l'$ values, around $l' = 50$. This observation suggests that a library that is too extensive can result in overfitting when $l'$ is increased beyond the intermediate range.

\begin{figure}[!]
	\centering
	\subfloat{\includegraphics[clip,width=0.35\columnwidth]{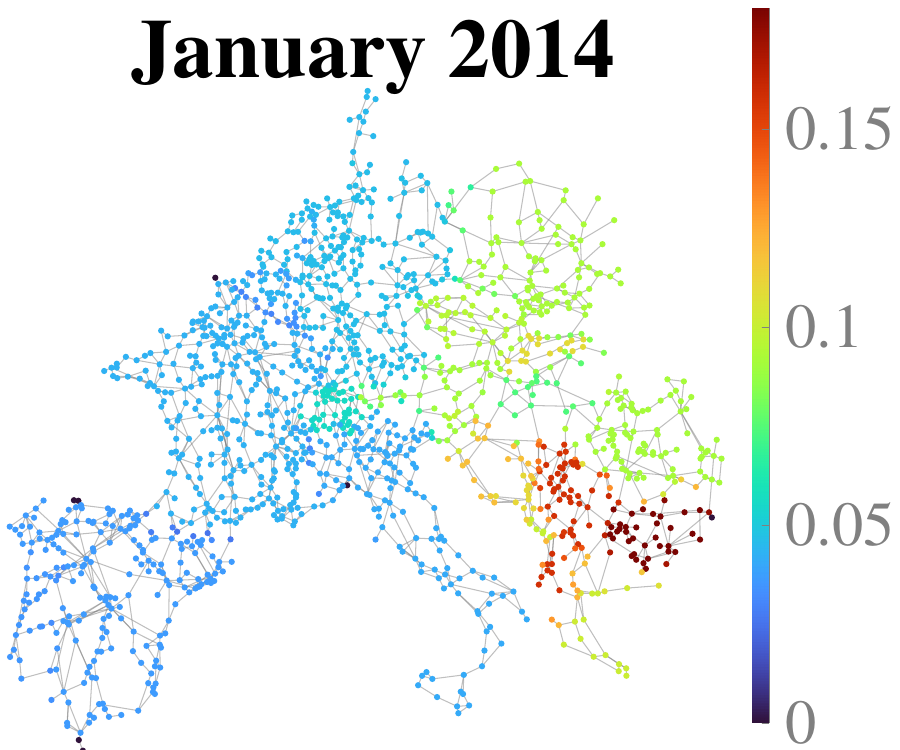}}    \hspace{20pt}
	\subfloat{\includegraphics[clip,width=0.35\columnwidth]{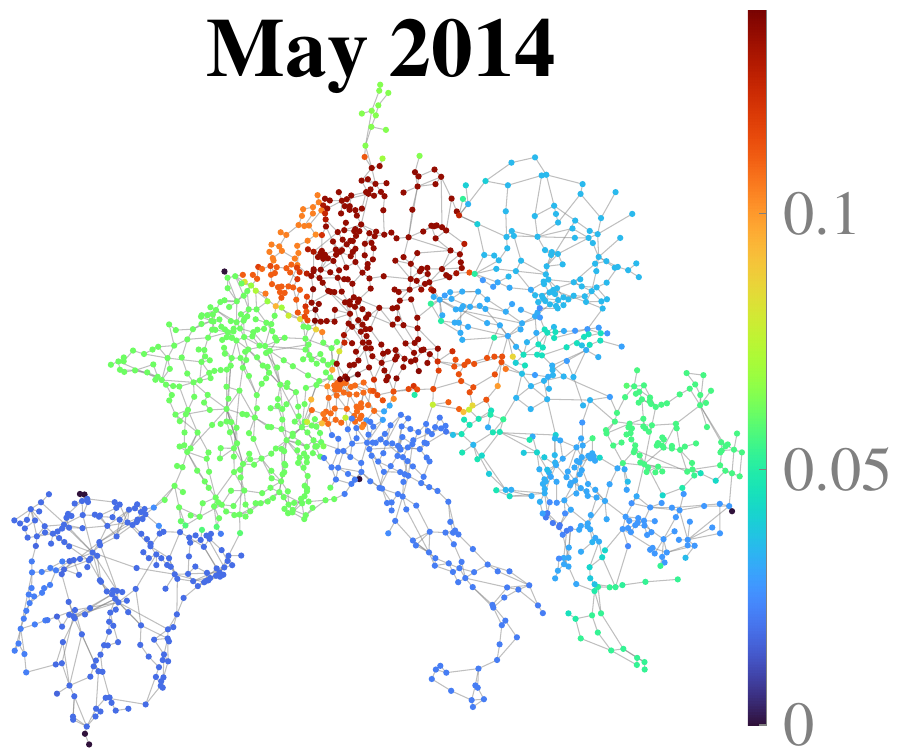}} \\
	\subfloat{\includegraphics[clip,width=0.35\columnwidth]{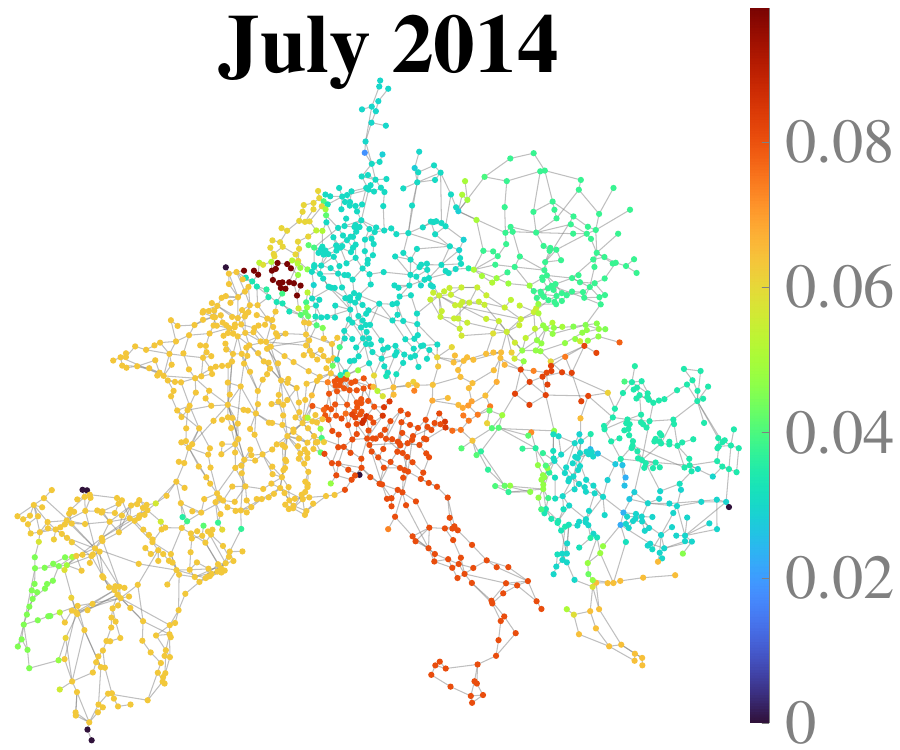}} \hspace{20pt}
	\subfloat{\includegraphics[clip,width=0.35\columnwidth]{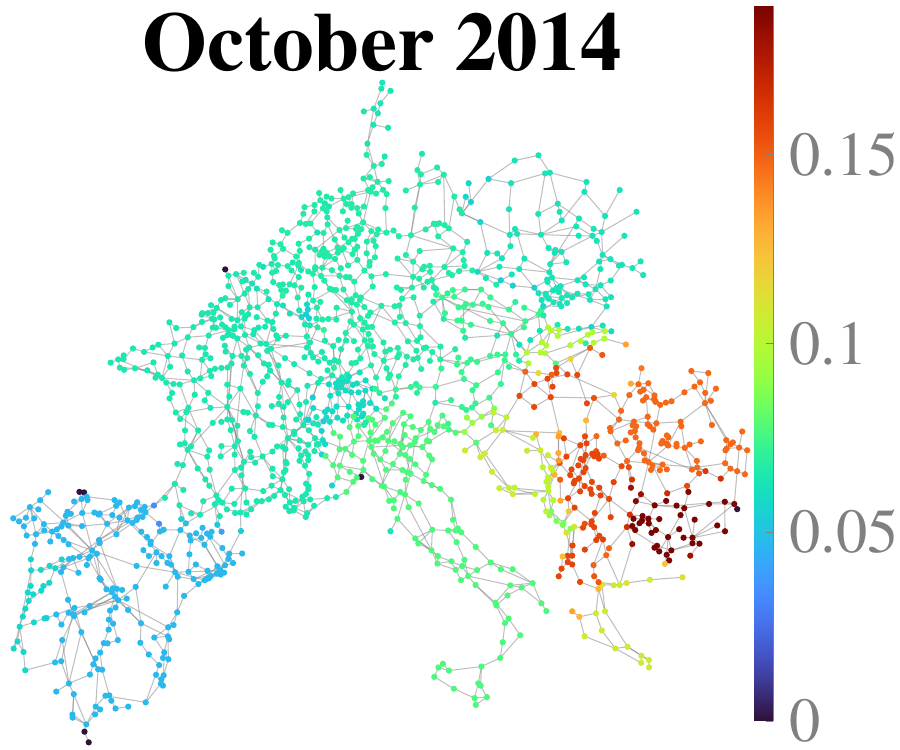}} 
	\caption{RMSE for different months with $Q=168 hr$ and $l'=50$.}
	\label{fig:RMSE_graph} 
\end{figure}

Figure \ref{fig:RMSE_graph} shows the RMSE values for different stations and different months of 2014, when $Q=168$ and $l'=50$. We apply these parameters to various Koopman models for different clusters and months. The four months January, May, July, and October in Figure \ref{fig:RMSE_graph} exemplify the load forecasting error in winter, spring, summer, and fall, respectively. The results demonstrate that the lowest forecasting errors occur during the winter and summer months, while errors are generally larger during spring and fall.
These seasonal variations in forecasting accuracy highlight the influence of various factors on the load dynamics. For example, the exceptionally high errors observed in April 2014 can be attributed to public holidays. Such observations underscore the complex interplay between seasonal changes and associated factors like weather patterns, social behaviors, and holiday schedules. These confounding influences can significantly impact load patterns and, consequently, forecasting accuracy. Future refinements of the model could potentially incorporate these factors to further improve forecasting performance across all seasons.

\begin{figure}[!]
	\centering
	\subfloat{\includegraphics[clip,width=0.6\columnwidth]{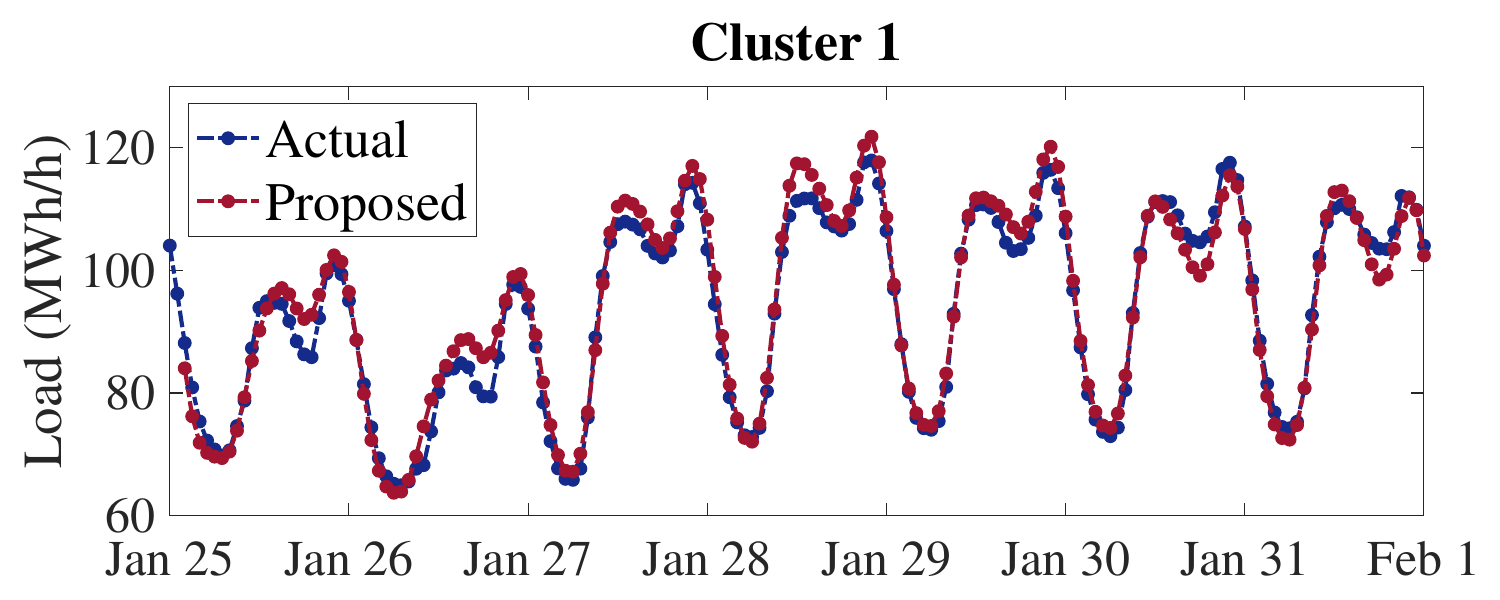}} 
	\caption{Example of forecasted load in last week of January 2014}
	\label{fig:ForecastJan2014} 
\end{figure}

Figure \ref{fig:ForecastJan2014} further demonstrates the performance of the proposed approach by showcasing examples of load forecasting in January 2014. For consistency across our analysis, we use the same parameters for different Koopman models across various clusters and months in Figures \ref{fig:RMSE_graph} and \ref{fig:ForecastJan2014}. Specifically, we set the number of delays $Q=168$, the number of Koopman eigenfunctions $l'=50$, and the regularization parameter $\epsilon=10^{-9}$.

\subsection{Comparison with deep learning}
\begin{figure}[!]
	\centering
	\subfloat{\includegraphics[clip,width=0.45\columnwidth]{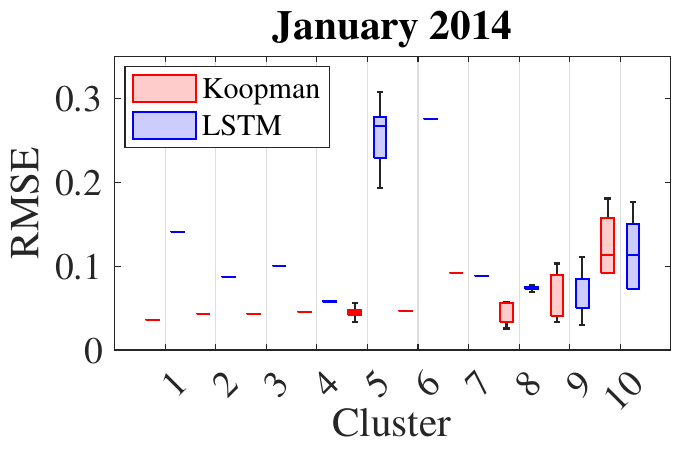}} \hspace{20pt}
	\subfloat{\includegraphics[clip,width=0.45\columnwidth]{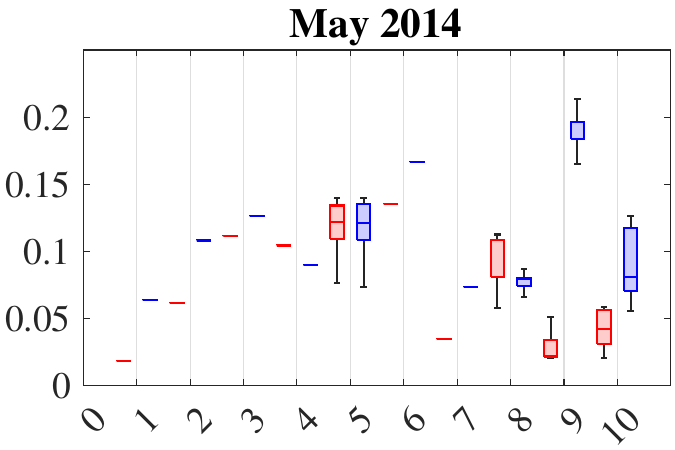}} \\
	\subfloat{\includegraphics[clip,width=0.45\columnwidth]{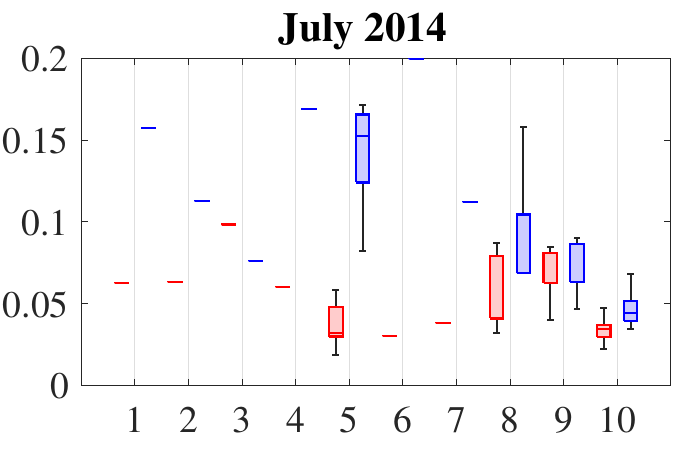}}     \hspace{20pt}
    \subfloat{\includegraphics[clip,width=0.45\columnwidth]{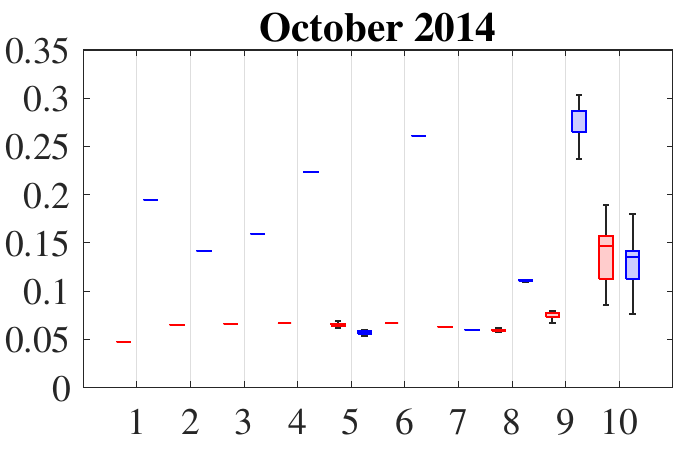}}
	\caption{Comparison of RMSE values between the Koopman-based approach and the LSTM model for different clusters across various months in 2014. Each subplot represents the last week of the respective month, showcasing seasonal variations in forecasting accuracy. Lower RMSE values indicate better performance.}
	\label{fig:RMSE_clu_compare} 
\end{figure}

Our approach unfolds and decomposes the data manifold structure to investigate system dynamics, providing a clear representation and ample room for interpretation. This method contrasts with conventional approaches that often treat the underlying system as a black box.
We compare our method to LSTM, a state-of-the-art approach for load forecasting \cite{Marino2016LSTM, Kong2019LSTM, Peng2021flexible, Zhu2022LSTM}. Figure \ref{fig:RMSE_clu_compare} compares RMSE values for different clusters and months of 2014, representing seasonal variations. For our Koopman-based approach, we used $l'=50$ eigenfunctions with a delay of $Q=168$.
For LSTM, we implemented a carefully optimized architecture to ensure a fair comparison. The LSTM used a two-layer architecture with 100 hidden units per layer and $\tanh$ activation function. We employed the Adam optimizer with mean squared error (MSE) loss function, a batch size of 64, and 10000 epochs. A dropout rate of 0.2 was applied after each LSTM layer. These parameters were chosen to balance model complexity, training time, and performance while minimizing overfitting.

Both methods predicted short-term forecasts for the final week of each month, using one week of training data. To ensure fairness, we designed a PHATE cluster-based LSTM, mirroring our approach with the Koopman method.
Figure \ref{fig:RMSE_clu_compare} shows lower forecasting errors for the Koopman approach across all clusters and seasons. 
Despite our careful design of the LSTM to balance efficiency and performance without unnecessary complexity, it still suffers from lack of interpretability and demands more computational resources, requiring 36 times more computation time than our Koopman approach for forecasting a month. 

	
\subsection{Modal noise and innovation noise}
\begin{figure}[!]
	\centering
	\subfloat{\includegraphics[clip,width=0.25\columnwidth]{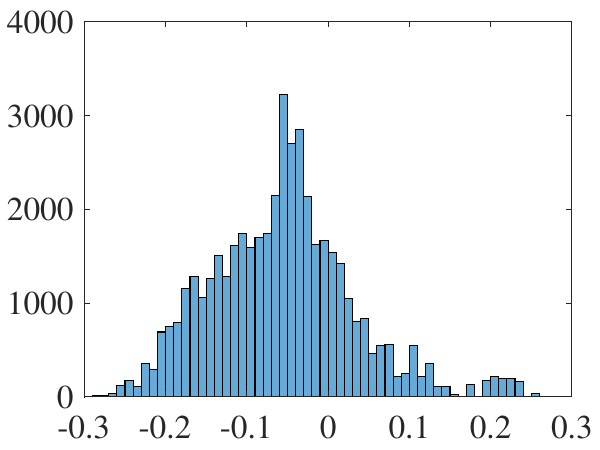}} \subfloat{\includegraphics[clip,width=0.25\columnwidth]{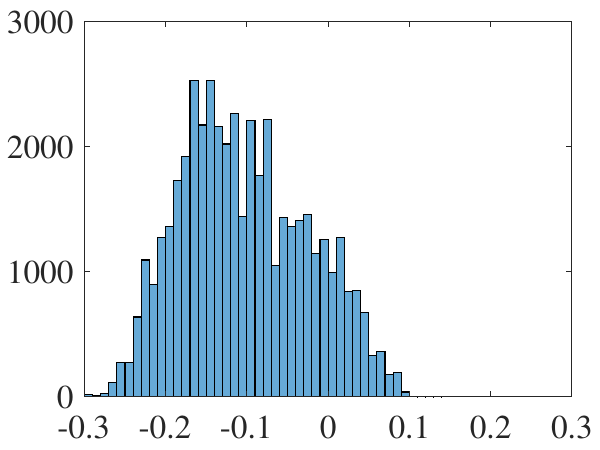}}
	\subfloat{\includegraphics[clip,width=0.25\columnwidth]{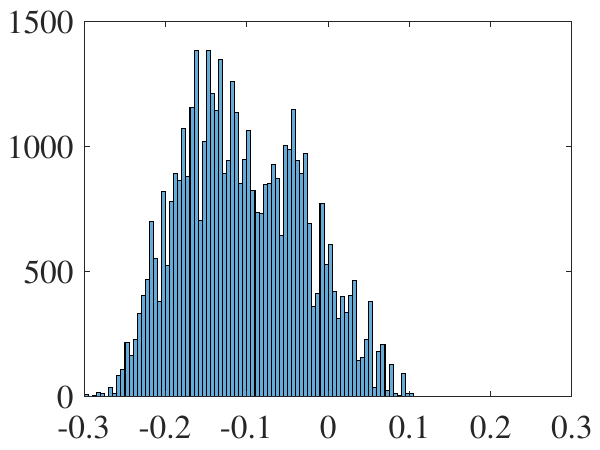}}
	\subfloat{\includegraphics[clip,width=0.25\columnwidth]{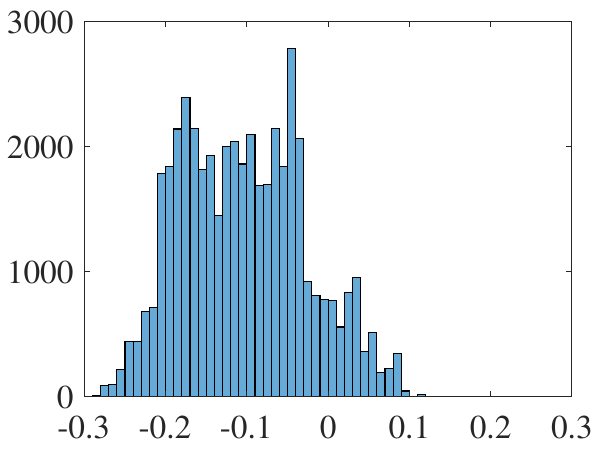}} \\
	\subfloat{\includegraphics[clip,width=0.25\columnwidth]{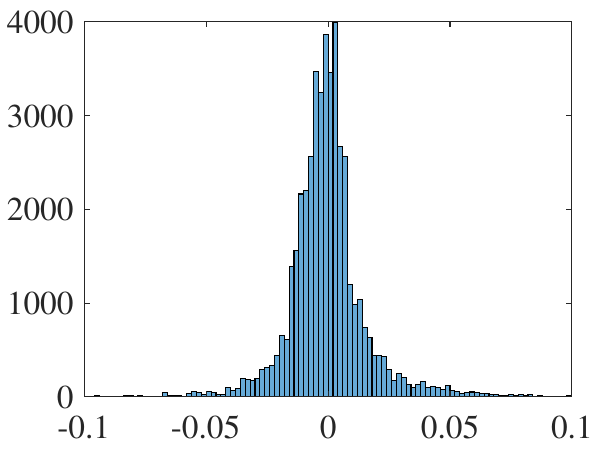}} \subfloat{\includegraphics[clip,width=0.25\columnwidth]{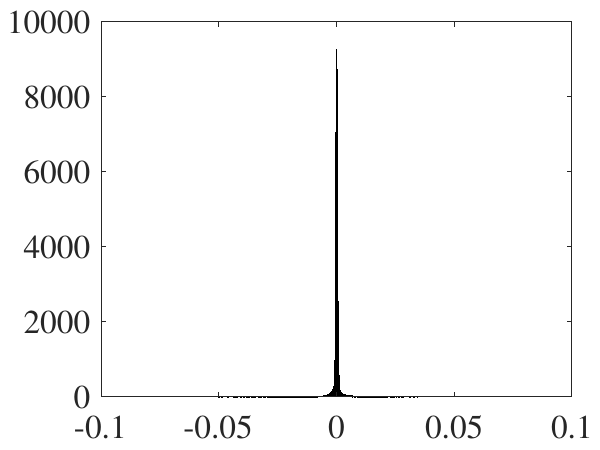}}
	\subfloat{\includegraphics[clip,width=0.25\columnwidth]{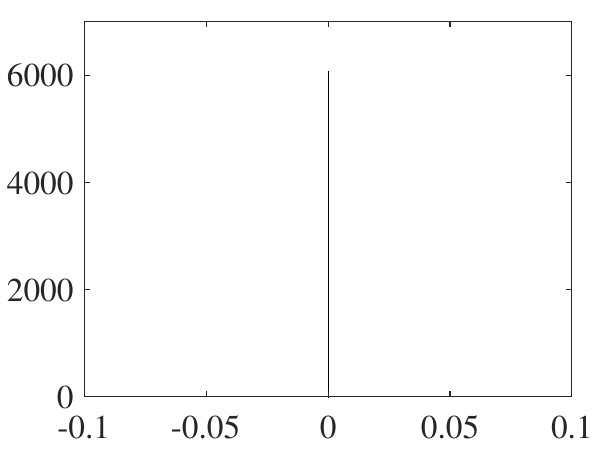}}
	\subfloat{\includegraphics[clip,width=0.25\columnwidth]{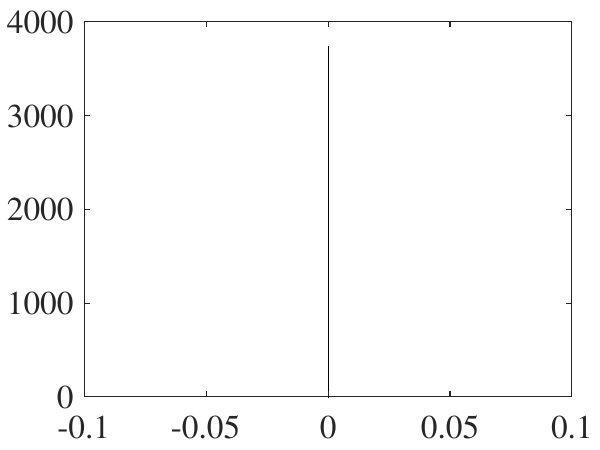}}
	\caption{Original data (without added noise): Modal (upper row)  and innovation (lower row) noise for load forecasting of all stations during 168 hours of last week of January 2014. Ordered from left, the columns show the results for $l'=5, 10, 20, 50$ Koopman eigenfunctions.}
	\label{fig:noise0} 
\end{figure}

\begin{figure}[!]
	\centering
	\subfloat{\includegraphics[clip,width=0.25\columnwidth]{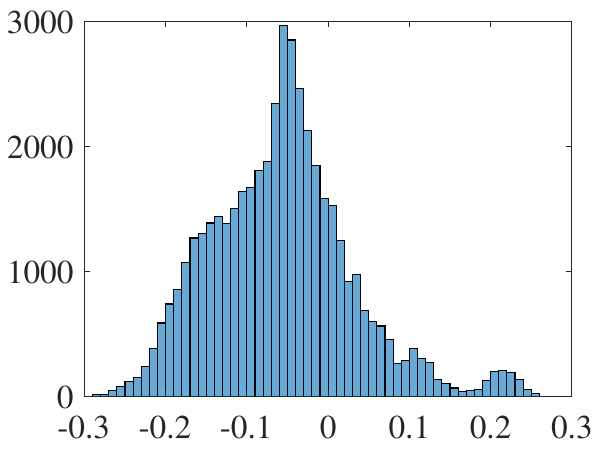}} \subfloat{\includegraphics[clip,width=0.25\columnwidth]{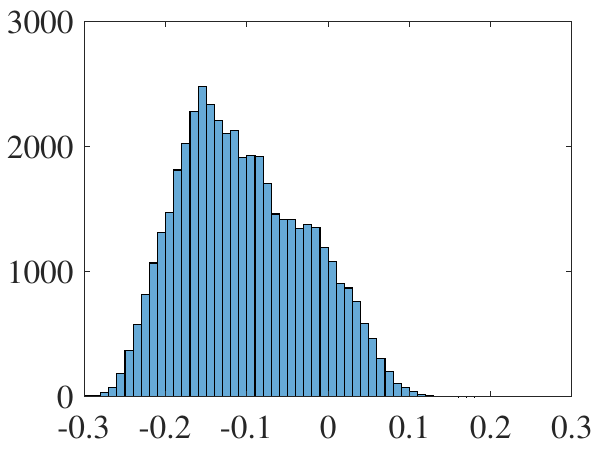}}
	\subfloat{\includegraphics[clip,width=0.25\columnwidth]{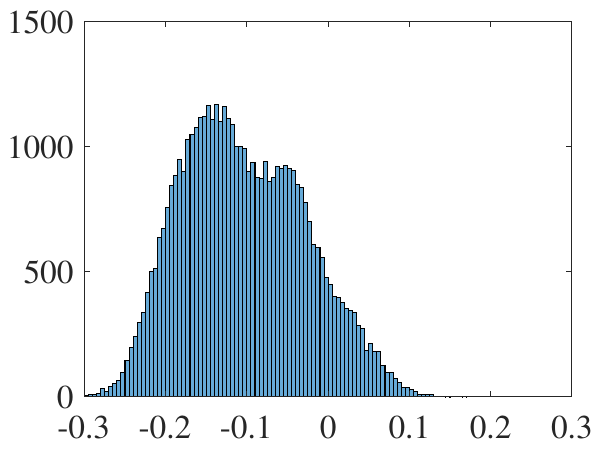}}
	\subfloat{\includegraphics[clip,width=0.25\columnwidth]{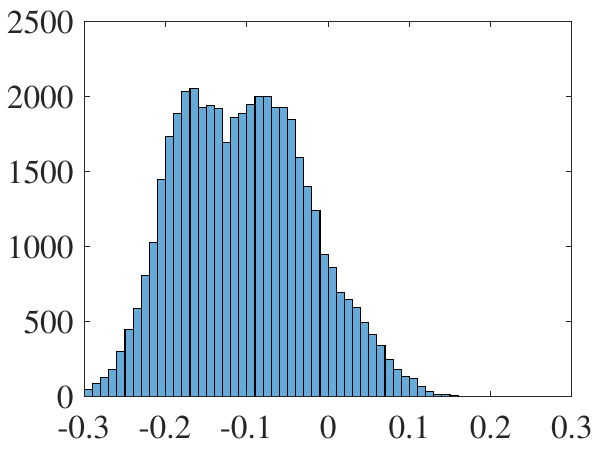}} \\
	\subfloat{\includegraphics[clip,width=0.25\columnwidth]{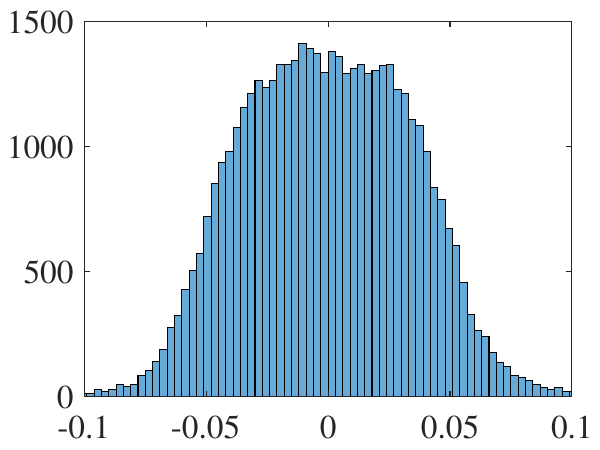}} \subfloat{\includegraphics[clip,width=0.25\columnwidth]{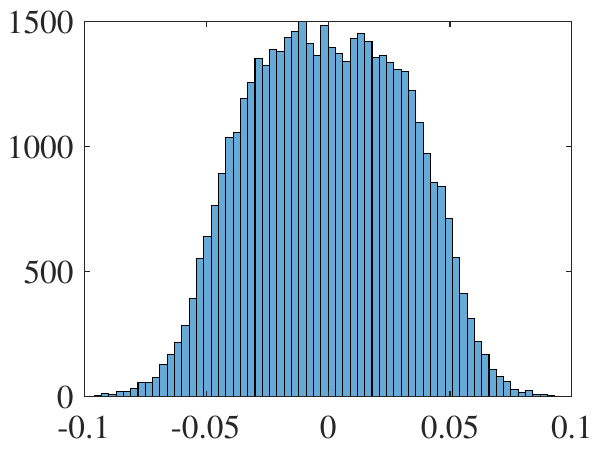}}
	\subfloat{\includegraphics[clip,width=0.25\columnwidth]{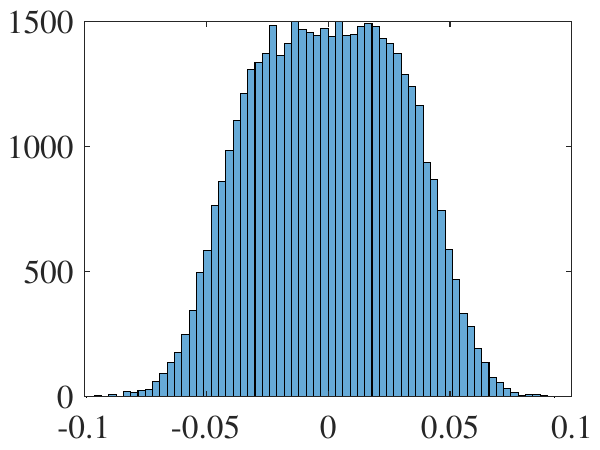}}
	\subfloat{\includegraphics[clip,width=0.25\columnwidth]{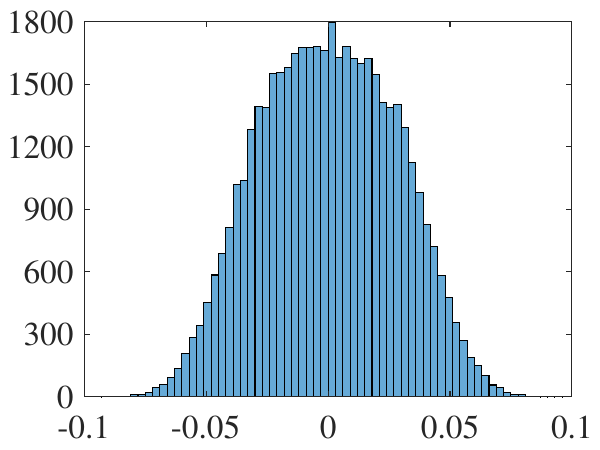}}
	\caption{Modal and innovation noise analysis with 10\% added measurement noise for load forecasting across all stations during the last week of January 2014 (168 hours). Upper row: Modal noise. Lower row: Innovation noise. Columns from left to right show results for $l'=5, 10, 20, 50$ Koopman eigenfunctions, demonstrating the effect of increasing the number of eigenfunctions on noise decomposition.}
	\label{fig:noise1} 
\end{figure}

By expressing the load dynamics using Koopman eigenfunctions, we can differentiate between various sources of forecasting errors and detect modal and innovation noises \cite{Mezic2022noise}. Due to the truncation of the infinite-dimensional Koopman operator, the finite-dimensional expansion $C\psi(x)$ introduces a residual $r$ in relation to the actual load data $x$, such that $x(t)=C\psi(t)+r(t)$. We split this residual $r$ into modal noise $\eta_m$ and innovation noise $\eta_i$ as follows:
\begin{equation}\label{eq:noise}
\begin{gathered}
\eta_m(t)=CC^\dagger r(t) \
\eta_i(t)=(I-CC^\dagger)r(t)
\end{gathered}
\end{equation}
where $I$ is the identity matrix and $C^\dagger$ is the Moore-Penrose pseudo-inverse of $C$.
The modal noise $\eta_m$ lies in the plane of truncated Koopman eigenfunctions, while the innovation noise $\eta_i$ remains perpendicular to this plane. Adding more eigenfunctions extends the modal plane and reduces innovation noise. However, modal noise, caused by mismatch in the regression matrix $C$ between training and test data, cannot be reduced by adding more Koopman eigenfunctions. This characteristic allows us to determine the optimal number of Koopman eigenfunctions \cite{Mezic2022noise}.
We analyze modal and innovation noise in power grids for January 2014, considering cases with and without measurement noise. For each case, we compute the residuals for all 168 hours of the last week of January 2014 and for all stations.
Figure \ref{fig:noise0} shows the case without measurement noise (original data without added noise). The upper row exhibits significant modal noise, while the lower row's innovation noise decreases with more Koopman eigenfunctions. This suggests that a few Koopman eigenfunctions suffice to capture effective dynamics modes in power grids, with the residual remaining in this space. Improving forecasting would involve modifying matrix $C$ and accurately evaluating Koopman mode contributions in load dynamics evolution. This could be achieved by incorporating exogenous effects, such as weather and public holidays, to modify the contribution of Koopman eigenfunctions in the training data for the test data's load dynamics evolution.
Figure \ref{fig:noise1} presents the case with measurement noise, where we have manually added 10\% noise to the original data. Here, the innovation noise changes only slightly as we increase the number of Koopman eigenfunctions (lower row). The Koopman approach enables us to differentiate between error caused by regression mismatch (upper row) and that due to measurement noise (lower row). Since the added noise lacks dynamics, it does not fall within the Koopman eigenfunctions' span. Thus, the Koopman model distinguishes between low-energy dynamic components and high-energy noisy parts, making no attempt to capture noise dynamics.

\section{Conclusion}\label{sec:Conclusion}
In this paper, we have presented a novel data-driven method for identifying and forecasting load dynamics in power grids using the Koopman operator approach. Our method identifies a reduced-order model for load dynamics by extracting coherent spatiotemporal patterns based on Koopman mode decomposition. These patterns provide a linear representation of the dynamics, enabling spatiotemporal spectral analysis and the discovery of fundamental energy patterns across multiple time scales.
We implement a cluster-based Koopman approach for load forecasting over large-scale power grids, utilizing the PHATE algorithm to detect synchronized load profiles and construct efficient, cluster-specific forecasting models. This approach demonstrates superior computational efficiency and suitability for large-scale power grids when compared to state-of-the-art deep learning methods, specifically LSTM networks. Moreover, our method offers enhanced interpretability, allowing for the identification and analysis of modal and innovation noise components in the forecasting process.
Importantly, our approach bridges the gap between physics-based and data-driven methodologies. It enables the discovery of hidden dynamical features of interconnected subsystems and exogenous effects in a nonparametric setting. This characteristic makes our method particularly valuable for understanding and predicting the complex behaviors of large-scale power grids.

The results presented in this paper demonstrate the potential of operator-theoretic approaches in power system analysis and forecasting. They suggest that such methods can provide valuable insights into system dynamics while maintaining computational efficiency and interpretability. Future work could explore the incorporation of additional exogenous factors, such as weather patterns and scheduled events, to further improve forecasting accuracy. Additionally, the application of this method to other complex networked systems could yield valuable insights across various domains of engineering and science.
In conclusion, our Koopman operator-based approach offers a promising new direction for load forecasting in power grids, combining the strengths of data-driven analysis with physically interpretable results. As power systems continue to grow in complexity and scale, such methods may become increasingly crucial for effective management and optimization of these critical infrastructures.

\bibliographystyle{IEEEtran}
\bibliography{refs}


\end{document}